\RequirePackage{fix-cm}
\documentclass[twocolumn]{svjour3}          
\usepackage{natbib}
\smartqed  
\usepackage{graphicx}
\usepackage{amsmath}
\usepackage{amssymb}
\usepackage{color}
\usepackage[table]{xcolor}
\usepackage{hyperref}
\usepackage{epsfig}
\usepackage{epstopdf}
\usepackage{subfigure}
\usepackage{tabularx}
 \usepackage[normalem]{ulem}
\usepackage{algorithm}
\usepackage[noend]{algpseudocode}
\usepackage{footmisc}

\usepackage[section]{placeins}
\usepackage{booktabs}
\usepackage{textcomp}
\usepackage{gensymb}
\usepackage{multirow}

\usepackage{makecell}
\usepackage{hhline}
\usepackage{cellspace}
\usepackage{comment}

\usepackage{mathtools}

\makeatletter
\def\BState{\State\hskip-\ALG@thistlm}
\makeatother
\definecolor{myGreen}{HTML}{33FF00}
\definecolor{myRed}{HTML}{FF3030}
\definecolor{myGrey}{HTML}{AA5555}
\definecolor{myWhite}{HTML}{FFFFFF}
\definecolor{maroon}{cmyk}{0,0.87,0.68,0.32}
\definecolor{petr}{HTML}{5555FF}
\definecolor{josef}{HTML}{FF3030}

\newcommand\scalemath[2]{\scalebox{#1}{\mbox{\ensuremath{\displaystyle #2}}}}

\newcommand\Set[2]{\{\,#1\mid#2\,\}}

\DeclareMathOperator{\cps}{CPS}
\DeclareMathOperator{\self}{Self6D}

\DeclareMathOperator{\avg}{avg}

\newcommand{\etal}[0]{\textit{et al.}}
\newcommand{\ie}[0]{\textit{i.e.}}
\newcommand{\eg}[0]{\textit{e.g.}}
\newcommand{\vs}[0]{\textit{vs.}}

\newcommand{\loss}[0]{\mathcal{L}}
\begin{document}

\title{CPS++: Improving Class-level 6D Pose and Shape Estimation From Monocular Images With Self-Supervised Learning
}

\titlerunning{CPS++: Improving Class-level 6D Pose and Shape Estimation}        

\author{Fabian Manhardt \and Gu Wang \and Benjamin Busam \and Manuel Nickel \and Sven Meier \and Luca Minciullo \and Xiangyang Ji \and Nassir Navab }

\authorrunning{F. Manhardt~\etal} 

\institute{Fabian Manhardt \at
           Technical University of Munich \\
           \email{fabian.manhardt@tum.de} 
           \and
           Gu Wang \at
           Tsinghua University 
           \and
           Benjamin Busam \at
           Technical University of Munich
           \and
           Manuel Nickel \at
           Technical University of Munich
           \and 
           Sven Meier \at
           Toyota Motor Europe
           \and 
           Luca Minciullo \at
           Toyota Motor Europe
           \and 
           Xiangyang Ji \at
           Tsinghua University 
           \and
           Nassir Navab \at
           Technical University of Munich
}
\vspace{-5mm}
\date{ }

\maketitle

\begin{abstract} 
Contemporary monocular 6D pose estimation methods can only cope with a handful of object instances. This naturally hampers possible applications as, for instance, robots seamlessly integrated in everyday processes necessarily require the ability to work with hundreds of different objects. To tackle this problem of immanent practical relevance, we propose a novel method for class-level monocular 6D pose estimation, coupled with metric shape retrieval. Unfortunately, acquiring adequate annotations is very time-consuming and labor intensive. This is especially true for class-level 6D pose estimation, as one is required to create a highly detailed reconstruction for all objects and then annotate each object and scene using these models. To overcome this shortcoming, we additionally propose the idea of synthetic-to-real domain transfer for class-level 6D poses by means of self-supervised learning, which removes the burden of collecting numerous manual annotations. In essence, after training our proposed method fully supervised with synthetic data, we leverage recent advances in differentiable rendering to self-supervise the model with unannotated real RGB-D data to improve latter inference. We experimentally demonstrate that we can retrieve precise 6D poses and metric shapes from a single RGB image.
\keywords{Class-level 6D Pose Estimation \and Self-supervised Learning \and Domain Adaptation}
\end{abstract}

\section{Introduction}

\begin{figure*}[t!]
    \centering
    \includegraphics[width=\linewidth]{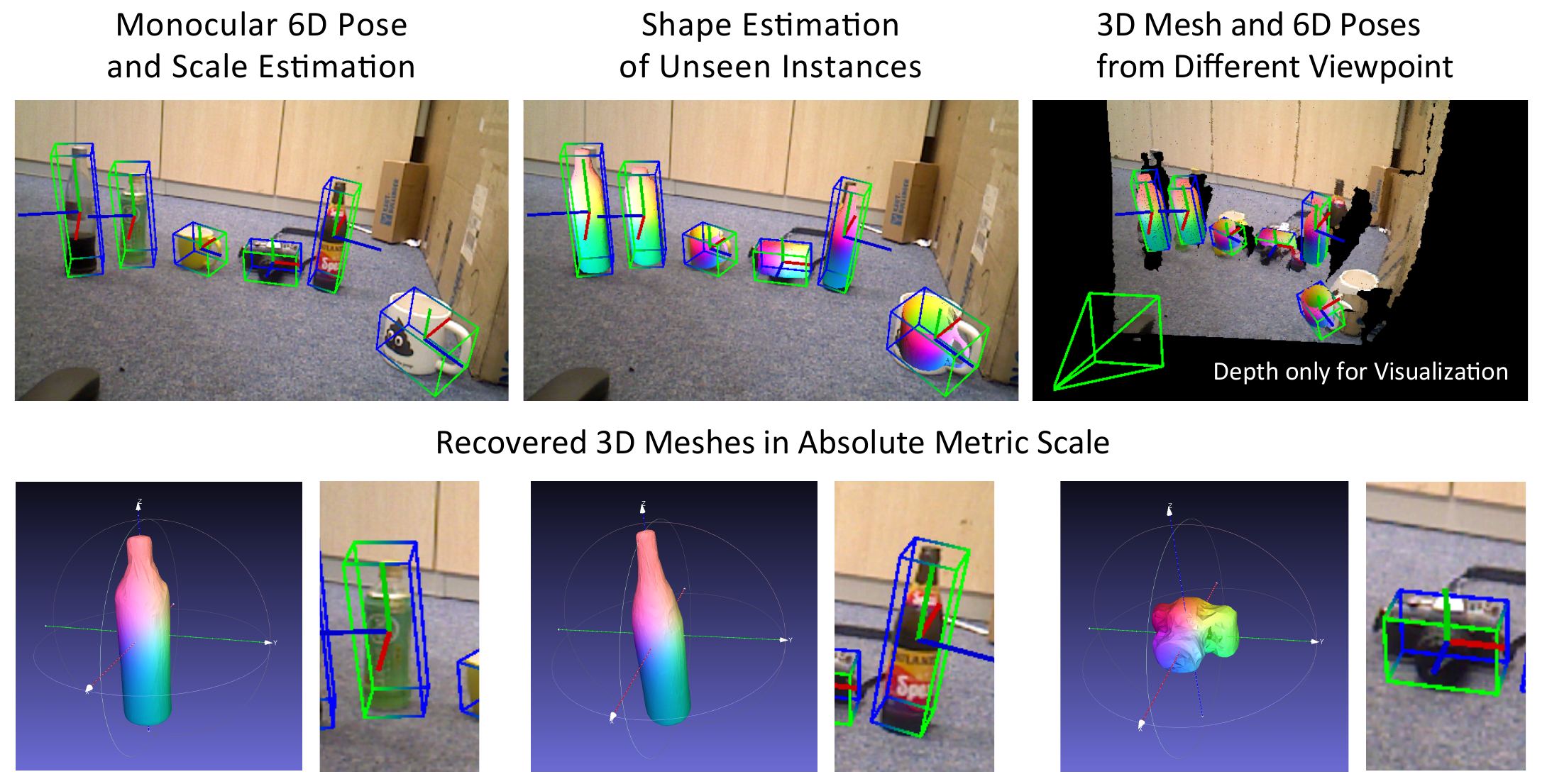}
    \caption{From a monocular image, we detect objects unseen during training from a particular class and estimate their 3D geometric properties such as 6D pose and metric scale (top left). Moreover, we infer the corresponding 3D shape for each detection on the bottom and rendered into the scene on the top center. To show that that the presented method is capable of inferring all parameters in correct 3D scale, we render our results from a different viewpoint (top right).\vspace{-1em}}
    \label{fig:teaser}
\end{figure*}

The field of 2D object detection has made huge leaps forward with the advent of deep learning. Current 2D object detectors can robustly detect more than a hundred different object classes in real-time \citep{liu2016ssd,tian2019fcos,wu2020recent}. Progress in the field of 6D pose estimation, however, is still limited due to the higher complexity of this task and the projective nature of images. 
 
Pioneering work has significantly improved the quality of object-specific 6D pose estimation \citep{kehl2017ssd,rad2017bb8}. However, little research is currently devoted to developing methods which are agnostic to the object type \citep{wang2019normalized}.
In fact, existing methods typically train separate networks for each object instance, which is slow and inflexible \citep{Sundermeyer_2018_ECCV,peng2019pvnet,park2019pix2pose}.

A few approaches have recently extended 3D object detection to object classes. These methods typically focus on automotive scenarios for detection and estimation of 3D bounding boxes of vehicles and pedestrians in outdoor environments where shape and pose variability is naturally bound \citep{chen2016monocular,chen2017multi}. Moreover, these works usually focus on large objects, leverage stereo \citep{li2019stereo} or lidar \citep{ku2018joint}, and limit the degrees-of-freedom for pose \citep{Xu2018}. Assuming all objects rest on the ground plane, many applications in this field restrict the output to 3D bounding boxes, whose only degree of freedom is the orientation of the vehicle which reduces the pose problem to an estimation of only 4 degrees-of-freedom \citep{chen2016monocular,chen2017multi}. A handful approaches for category level 3D object detection in indoor environments have also been proposed \citep{song2016deep, nie2020total3dunderstanding}.

In the field of robotics, where two of the main applications are grasping and manipulation, the 3D bounding boxes is oftentimes an insufficent parameterization. This task also imposes different constraints: objects are often physically and visually small, depth is not always present or incomplete for certain objects and the ground-plane assumption is often unrealistic (\textit{e.g.} in Fig~\ref{fig:pose_comparison_real}: the camera on the top exhibits in-plane rotation and the can object on bottom stands on a different level than all other objects). Thus, 6D pose estimation is necessary. Due to the increase in degrees of freedom, this task becomes more challenging. Recently, \citet{wang2019normalized} introduced the first method for class-level 6D pose estimation. While their innovative work enabled instance-agnostic 6d pose measurements, an additional depth map was indispensable. \citet{chen2020learning} estimate the full 6D pose paired with the object's shape in the form of a point cloud. Nonetheless, similar to \citep{wang2019normalized}, \citet{chen2020learning} also expect the presence of the depth map. Moreover, both methods employ labeled real data to be capable of successfully training their models.

Annotating the 6D object pose, however, requires additional special hardware \citep{garon2018framework}, is kno\-wn to be extremely time-consuming \citep{xiang2017posecnn} as well as error prone \citep{hodan2019photorealistic,tremblay2018falling}. This is especially true when dealing with object classes, as each object instance requires a high-quality 3D scan, which then needs to be manually fitted into all the training images in a very tedious process. Furthermore, it does not scale well when introducing new classes as the whole process needs to be repeated. On the other hand, it is relatively easy to record a large amount of RGB images without annotations even with consumer hardware. Self-supervised learning is a new research direction which focuses on learning despite the lack of appropriate annotations \citep{godard2017unsupervised,kocabas2019self}.

As illustrated in Fig.~\ref{fig:teaser}, we propose $\cps$, a novel method for monocular \textbf{C}lass-level 6D \textbf{P}ose and metric \textbf{S}hape estimation, enabling new applications in augmented reality, and robotic manipulation. To ensure wide applicability of our method, we regress shape and pose parameters from a single RGB input image. This is particularly difficult due to the inherent ambiguities of 3D inference from 2D images. Inspired by \citep{Manhardt2019,Simonelli2019}, we introduce a novel 3D lifting module which directly aligns a predicted point cloud for each detection in 3D camera space. In contrast to other methods \citep{Manhardt2019} which predict shape in a dedicated branch, trai\-ned independently from the pose estimation network, we back-propagate the final alignment through the entire network. Our method is thus trainable in an end-to-end manner and directly optimizes for the best alignment in 3D. Since we need pose and shape annotations to train our object detector while avoiding extra labeling effort, we rely fully on synthetic data. However, this introduces a large domain gap against the real world. Inspired by $\self$ \citep{wang2020self6d} and recent trends in self-supervised learning \citep{godard2017unsupervised,kocabas2019self}, we thus want to train our pose estimator on such unsupervised samples. To this end, we tailor $\self$ towards the problem of class-level 6D pose estimation in order to transfer the knowledge from the synthetic to the real environment with a self-supervision loss.

In summary, we make the following contributions. To the best of our knowledge, i) we are the first to introduce the task of monocular 6D pose paired with metric shape estimation and ii) propose $\cps$, a novel method which directly aligns the final outcome in 3D setting a new state of the art for pose accuracy while it is also able to estimate object shapes. iii) We additionally introduce a self-supervised extension of $\cps$ to bridge the synthetic-to-real domain gap that also works with object classes; we dub it $\cps$++. To this end, iv) we also collected over 30k real RGB-D samples, which we made publicly available. Finally, v) we also introduce a new metric for joint shape and pose estimation, which we call \textit{Average Distance of Predicted Point Sets}.
\section{Related Work}

We first introduce essential recent works in monocular instance-level 6D object pose estimation. We then discuss first approaches to class-level 6D object pose estimation. Since most works for monocular class-level 3D object detection are found in the autonomous driving community, we also outline the most relevant works there. We also take a look at recent trends in 3D shape recovery. Finally, we discuss current developments in neural rendering and review some first attempts at self-supervised learning for 6D pose.

\subsection{Monocular 6D Object Pose Estimation}
Traditionally, object pose estimation approaches rely on local image features \citep{lowe1999object,Romea-2011-7355} or template matching \citep{hinterstoisser2012gradient}. With the advent of consumer RGB-D cameras, the focus mov\-ed more towards conducting object pose estimation from RGB-D data. While some works again propose to utilize template matching \citep{Hinterstoisser2012}, others leverage point pair features \citep{vidal2018method} or rely on learning-based methods \citep{brachmann2014learning,krull2015learning} in order to predict the 6D pose.

Nonetheless, depth data also comes oftentimes with limitations such as restricted field of view or high power consumption. Recently, CNN-based methods have de\-monstrated promising results for the task of monocular 6D pose estimation \citep{hodan2018bop}.

A few methods directly regress the 6D pose. For instance, \citep{xiang2017posecnn,li2019deepim} learn to estimate poses through the minimization of a point matching loss. In contrast, \citet{kehl2017ssd} discretize the pose space and classifies viewpoint and in-plane rotation. \citet{manhardt2019ambiguity} adopts \citep{kehl2017ssd} to implicitly handle ambiguities via multiple hypotheses. A different line of works learn a latent embedding for the discretized pose space and recover 6D poses using codebook matching \citep{sundermeyer2018implicit,sundermeyer2020multi}.

Another popular branch is to establish 2D-3D correspondences and solve the 6D pose using P$n$P with RANSAC. \citet{rad2017bb8,tekin2018real} propose to estimate the 2D projections of a fixed set of 3D keypoints in image space. Similarly, \citet{hu2019segpose,peng2019pvnet} further extend this idea by employing segmentation paired with voting to improve robustness. 
In contrast, \citet{zakharov2019dpod,li2019cdpn,park2019pix2pose,hodan2020epos} predict object coordinates in order to establish dense 2D-3D correspondences, rather than sparse ones. 

\subsection{Beyond Instance-Level 6D Pose Estimation} 
\citet{wang2019normalized} recently proposed the first method for class-level object detection and 6D pose estimation. \citet{wang2019normalized} predict a 2D map representing the projection of the Normalized Object Coordinate Space (NOCS). The NOCS is a 3D space within a unit cube. All objects within a categories are normalized to lie within the NOCS, allowing to handle even unseen object instances of the corresponding category. This 2D NOCS map is then backprojected, using the associated depth map, to establish 3D-3D correspondences. Leveraging these correspondences together with the Umeyama algorithm \citep{umeyama1991least} enables the estimation of both 6D pose and scale. \citet{chen2020learning} instead propose to conduct class-level object pose and size estimation with a correspondence-free approach. They learn a canonical shape space for input RGB-D images with normalized shape and metric size based on a deep generative model before estimating the pose by comparing the pose-independent and pose-dependent features. \citet{park2020latentfusion} further propose a novel framework for 6D object pose estimation of fully unseen objects without any prior information. Nonetheless, this method require to compute gradients during inference which is slow and, additionally, assumes reference images in order to reconstruct the latent 3D object.
Notice that all these methods expect annotated real data and the presence of a depth image during inference. We instead do not need labeled real data and only use monocular data to predict 6D pose, object shape and metric size.

\subsection{Monocular Class-Level 3D Object Detection} 
Classical approaches rely on shape based classification with pose parametrization by 3D geometric primitives \citep{carr2012monocular}. The parametrization paradigm has been relaxed by \citet{chen2016monocular} who use multiple monocular cues such as shape, segmentation, location, and spatial context to instantiate 3D object proposals followed by a CNN-based scoring. \citet{kundu20183d} predict rotation and shape of cars employing a render-and-compare loss. 
\citet{Manhardt2019} introduce a 3D lifting loss which measures the misalignment of the 3D bounding box corners. 
In addition, they also learn a shape space for \emph{truncated sign distance functions} (TSDFs) using a 3D auto-encoder and train a sub-network to predict the latent representation for each detection. 
Nonetheless, these methods predict shape either only up to scale \citep{kundu20183d} or neglect it during optimization for pose and learn it at a later stage \citep{Manhardt2019}. 
However, \citet{chen2016monocular} show that shape can provide extra cues on the pose and should not be dissociated. \citet{Simonelli2019} similarly measure the 3D bounding box misalignment, however, compute the error for each pose parameter separately to improve stability during training. 
\citet{ku2019monocular} propose to leverage instance-centric 3D proposal and local shape reconstruction. \citet{ma2019accurate} first conduct monocular depth prediction to produce a pseudo lidar. Afterwards, they employ a PointNet architecture to obtain the objects’ poses and dimensions. \citet{ding2020learning} propose to employ depth-guided local convolutions instead of pseudo lidar to better process the predicted depth maps. Finally, \citet{chen2020monopair} attempt at improving monocular 3D object detection by considering mutual spatial relationships of objects.

Interestingly, almost all these methods assume all objects to be standing on the ground plane and only estimate one angle for the object's orientation with respect to the plane, thus, reducing pose to a problem with 4 degrees-of-freedom under additional constraints.

\subsection{Recent Trends in Rigid 3D Shape Recovery} 
\citet{groueix2018atlasnet} introduce AtlasNet, a network architecture built on top of PointNet \citep{qi2017pointnet}. Sampling points from 2D uv-maps, the network can reconstruct 3D shapes of arbitrary resolution. 
Also based on PointNet, \citet{yang2018foldingnet} propose a light-weight end-to-end trainable encoder-decoder architecture, that learns to deform a 2D grid into the 3D object surface of a point cloud.

\citet{wang2018pixel2mesh} propose Pixel2Mesh, which directly regresses 3D meshes from RGB. The network iteratively refines the geometry of an input 3D ellipse, using features extracted from a single RGB image employing a graph convolutional neural network. \citet{gkioxari2019mesh} introduce Mesh~RCNN, an extension of Mask~RCNN \citep{he2017mask}. This network estimates a voxel representation for objects, which can be refined by a series of graph convolution-based refinement steps. 

Finally, another recent trend is to make use of implicit functions for 3D shape recovery \citep{mescheder2019occupancy,Park_2019_CVPR, genova2020local,niemeyer2020differentiable,deng2020cvxnet}. Thereby, the 3D surface is represented by the continuous decision boundary of a deep neural network classifier. Once the networks are trained, the object surface can be extracted from the learnt boundary.

\subsection{Differentiable Rendering For 3D Meshes}
Most traditional rendering pipelines are usually not differentiable due to the rasterization step, as they rely on hard assignments of the closest triangle for each pixel \citep{nguyen2018rendernet}. 
Therefore, many works have recently been proposed to circumvent the hard assignment in order to re-establish the gradient flow \citep{kato2020differentiable}.

Early attempts try to approximate the gradients of pixels with respect to the mesh's vertices \citep{opendr_eccv14,kato2018renderer}. 
More recent works instead approximate the rasterization itself in order to obtain analytical gradients. 
For instance, \emph{SoftRas} conducts rendering by aggregating the probabilistic contributions of each mesh triangle in relation to the rendered pixels \citep{liu2019softras}. \emph{DIB-R} extends \emph{SoftRas} by considering foreground and background pixels independently \citep{chen2019learning_dibrenderer}. 
\citet{wang2020self6d} further adjust \emph{DIB-R} to conduct a real perspective projection and additionally render the associated depth map.

\subsection{Self-Supervised Learning For 6D Pose} 
While most works in literature used to either rely on Generative Adversial Networks \citep{bousmalis2017unsupervisedPixelda,lee2018diverse} or make use of domain randomization \citep{kehl2017ssd,zakharov2019deceptionnet} to avoid the need for real data with 6D pose annotations, a few methods recently proposed to instead harness ideas from self-supervised learning. 
In essence, self-supervised learning describes learning from unlabeled real data, where the supervision comes from the data itself, and has recently enabled a large number of applications in computer vision. Supervision is commonly achieved by enforcing different constraints such as consistencies from geometry, multiple views, or multiple modalities \citep{godard2017unsupervised,kocabas2019self,kolesnikov2019revisiting}. 

In the field of 6D pose, \citet{deng2020self} propose a self-labeling pipeline for RGB-D based 6D object pose estimation with an interactive robotic manipulator. In contrast, \citet{Zakharov_2020_CVPR} propose a curriculum learning strategy. They iteratively label the training data, then optimize these annotations using differentiable rendering and retrain the 3D object detector. However, the core of both 6D pose estimation modules is still trained fully-supervised using the self-labeled data.

In contrast, \citet{wang2020self6d} recently introduced $\self$, which directly learns pose from the raw data without any labeling. Given a trained 6D pose estimation network and unlabeled RGB-D data, \citet{wang2020self6d} enforce consistency between the query data and the predicted poses leveraging differentiable rendering. Essentially, using DIB-R \citep{chen2019learning_dibrenderer}, \citet{wang2020self6d} render an RGB and depth image which is then visually and geometrically aligned with the sensor input.
\section{Class-level Monocular Pose \& Metric Shape}

\begin{figure*}[t!]
    \centering
    \includegraphics[width=0.97\textwidth]{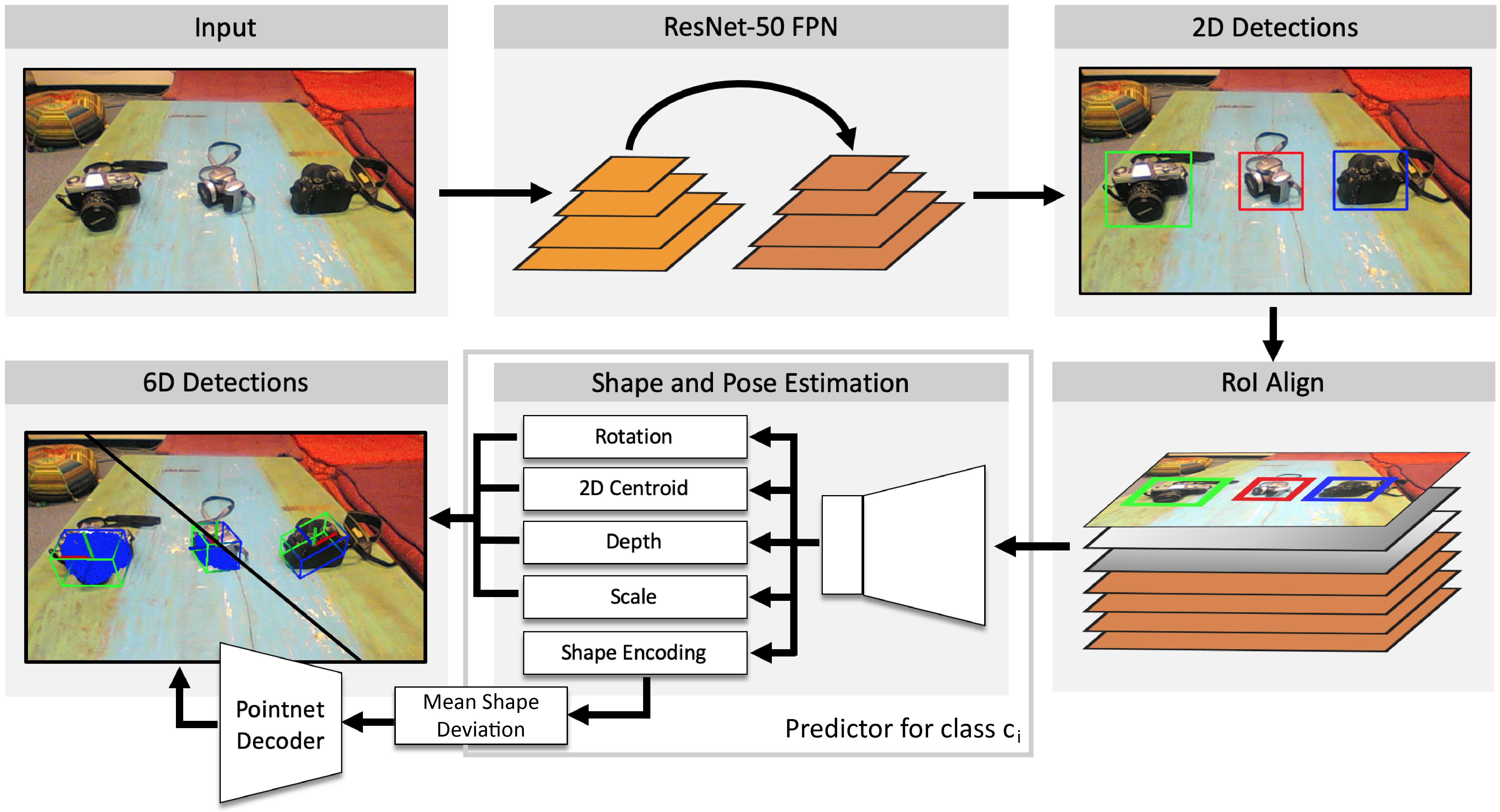}
    \caption{\textbf{Schematic overview.} We feed the input image to a RetinaNet to infer 2D detections. We then collect all detections for each class and send them to the associated lifter module, which predicts the 6D pose together with the scale and the shape encoding. Finally, we retrieve point clouds from AtlasNet.}
    \label{fig:method}
\end{figure*}

In this section, we introduce our method for estimating the 6d object pose and metric shape, represented as point cloud, from a single RGB image. We first describe how we learn an explainable shape space for each class using a PointNet \citep{nguyen2018rendernet} auto-encoder and then present our proposed architecture for 6D pose and metric shape. Finally, we depict our novel loss for aligning the extracted 6D pose and shape in 3D space and demonstrate how we conduct domain transfer from synthetic to real data.

\subsection{Learning an Explainable Shape Space}
A core novelty of our work lies in the joint estimation of the object's shape alongside its 6D pose from a single RGB image. Inspired by commonly used low-dimensional embeddings in the domain of shape estimation \citep{kundu20183d,manhardt2019roi}, we decided to employ a 32 dimensional latent space representation for each class. 
During inference, this enables the reconstruction of a 3D model by predicting only few shape parameters as opposed to a complete point cloud.

We employ AtlasNet \citep{groueix2018atlasnet} to learn a latent space representation of an object class~$c$. The network is based on PointNet \citep{qi2017pointnet} and takes as input a complete point cloud which it then encodes into a global shape descriptor. One can reconstruct a 3D shape by concatenating that descriptor with points sampled from a 2D uv-map and feeding the result to a decoder network. This approach decouples the number of predicted points from that in the original training shapes, thus enabling the reconstruction of shapes with arbitrary resolution. We decided to employ AtlasNet due to the fact that the triangles for meshing can be inferred easily from the employed 2D uv-map. This makes it particularly useful when rendering the predictions for our self-supervision. We train one AtlasNet network for each object class separately on a subset of point clouds $\mathcal{P}_c$ from ShapeNet \citep{chang2015shapenet}, each one learning a class specific distribution of valid shapes in latent space.

\subsection{Differentiable 6D Pose and Metric Shape}
Since our self-supervision requires the flow of gradients throughout the whole network, we cannot resort to any method which is based on establishing non-differentiable 2D-3D correspondences. Thus, we rely on a similar architecture as \citep{manhardt2019roi}. As illustrated in Fig.~\ref{fig:method}, our method is based on a two-stage approach similar to Faster R-CNN \citep{ren2015faster}. We first predict 2D regions of interest using RetinaNet with Focal Loss \citep{lin2017focal}. The object proposals are then processed by our pose and shape estimator. We employ a ResNet-50 backbone with an FPN structure. For each detected object, we apply RoIAlign \citep{he2017mask} to crop out regions of interest with size of $32\times32$. We also apply the RoIAlign operator on the input RGB image and the coordinate tensor \citep{liu2018intriguing}. 
Thus, the pose predictor is aware of the location of the crop and does not lose global context. We concatenate both RoIAlign outputs with the feature maps from FPN to compute the feature map $f$ for the given 2D detection.

\subsubsection{From 2D to 3D Detection} 
For each RoI, separate predictor networks branch off to infer: a 4D quaternion $q_a$ representing the 3D rotation in $SO\left(3\right)$, the 2D centroid $(x,y)$ as the projection of the 3D translation into the 2D image given camera matrix $K$, the distance $z$ of the detected object with respect to the camera, the metric size $(w, h, l)$ of the object, and the low-dimensional representation $e$ of the shape. In addition, we also predict the object mask $M_P$ as it plays a crucial role in out latter self-supervision.

\begin{figure}[t!]
    \centering
    \includegraphics[width=0.99\linewidth]{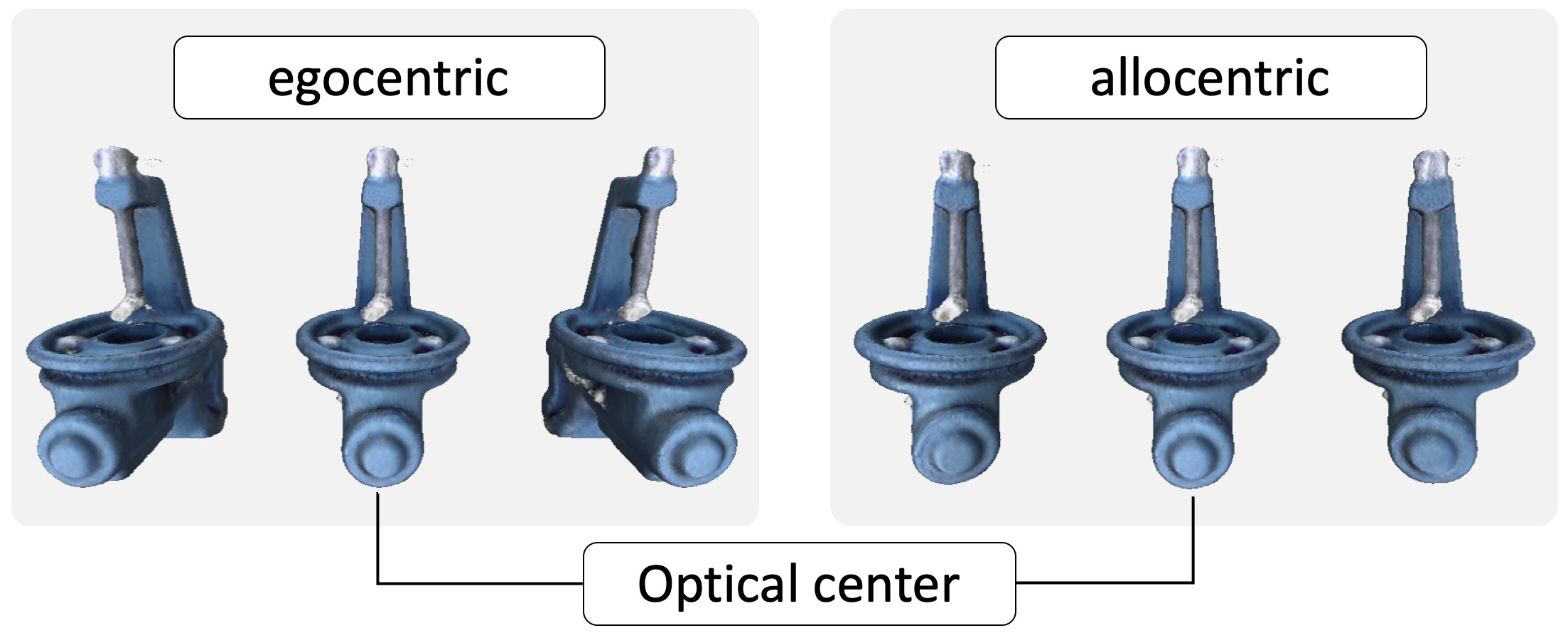}
    \caption{\textbf{Egocentric \vs allocentric rotation.} Under egocentric projection, a mere 3D translation of the object lateral to the image plane, leads to different object appearance. This is not the case under allocentric projection.}
    \label{fig:ego_allo}
\end{figure}

The final pose is obtained by back-projecting the 2D centroid with respect to the regressed depth and known camera matrix $K$ to compute the 3D translation $t = K^{-1} z\left(x, y, 1\right)^T$. Then, we use the translation to compute the egocentric rotation $q$ from the predicted allocentric rotation $q_a$. Since we deal with cropped RoIs, the allocentric representation is favored as it is viewpoint invariant under 3D translation of the object \citep{kundu20183d,mousavian20173d}. The difference is visualized in Fig.~\ref{fig:ego_allo}. Note that knowing the translation, one can easily convert from the allocentric to the egocentric representation.

Given the object’s estimated allocentric rotation $q_a$, the 2D projection $c$, and the camera matrix $K$, we first calculate the rotation $q_c$, between the camera principal axis $[0,0,1]^T$ and the ray through the object center projection $K^{-1}c$. Then we compute the rotation that takes vector $[0,0,1]^T$ to align with vector $K^{-1}c$ according to
\begin{equation}
    q_c \coloneqq [\cos{\frac{\alpha}{2}}, A(0) \cdot \sin{\frac{\alpha}{2}}, A(1) \cdot \sin{\frac{\alpha}{2}}, A(2) \cdot \sin{\frac{\alpha}{2}}]
\end{equation}
with $A = [0,0,1]^T \times K^{-1}c$ being the axis between the object centroid $K^{-1}c$ and the optical center ray $[0,0,1]^T$ and $\alpha = \arccos{(K^{-1}c})$ describing the angle between them. The final egocentric rotation is then computed according to $q = q_c \cdot q_a$.

Since features from the FPN stage are forwarded to the second stage, we only require very small lifting networks for pose ($q_a$, $(x,y)$, $z$) and shape ($e$, $(w, h, l)$). Thus, we can easily afford to use separate lifting modules for each object class. In practice, each detected object is forwarded to its corresponding lifting module given the estimated class label. Therefore, poses and shapes from different classes do not interfere during optimization. For each lifter we first apply two 2D convolutions with batchnorm before diverging into separate branches for pose and shape. For each branch, we employ another two 2D convolutions with batchnorm followed by a fully-connected layer to predict the final parameters.

\subsubsection{Retrieving 3D Shape}
Using the AtlasNet encoder $\mathcal{E}_c$, we compute the bound feature
\begin{equation}
 \mathcal{S}_c \coloneqq \Set{\mathcal{E}_c(p)}{p \in \mathcal{P}_{c}} \subset \left[-1,1\right]^{32},
\end{equation}
which is the set of all latent space representations of the training shapes.
From $\mathcal{S}_c$, we then calculate a per-class mean latent shape
\begin{equation}
    m_c \coloneqq \frac{1}{|\mathcal{S}_c|}\sum_{s_c \in \mathcal{S}_c} s_c.
    \label{mean_shape}
\end{equation}
Let us denote the shape prediction branch as $\mathcal{F}_{Shape}(f)$, a non-linear function that outputs a class-specific latent shape vector for the feature map $f$ from the given RoI. Then, instead of forcing $\mathcal{F}_{Shape}(f)$ to predict absolute shape vectors~$e$, we let it infer a simple offset from $m_c$, such that $e \coloneqq m_c + \mathcal{F}_{Shape}(f)$.
Finally, the AtlasNet decoder network reconstructs a 3-dimensional point cloud, \ie~$p(f) \coloneqq \mathcal{D}_c(m_c + \mathcal{F}_{Shape}(f))= \mathcal{D}_c(e)$.

To encourage the latent shape predictions of $\mathcal{F}_{Shape}$ to stay inside of the learned shape distribution, we employ a special regularization loss. Assuming the shape encodings of the per-class training span a convex shape space $Conv({S}_c)$, we punish the network for any predicted $e\not\in Conv(\mathcal{S}_c)$ and project them onto $\partial{\mathcal{S}_c}$, the boundary of $Conv(\mathcal{S}_c)$. In practice, we detect all $e\not\in Conv(\mathcal{S}_c)$ as
\begin{equation}
         I(e | \mathcal{S}_c) = 
\begin{cases}
    0,  & \text{if} \min\limits_{\substack{s_{c,i}, s_{c,j} \in \mathcal{S}_{c} \\ i\neq j}} (e - s_{c,i})^{T} (e - s_{c,j}) \leq 0 \\
    1,              & \text{otherwise}.
\end{cases}
\end{equation} 
where $I(e|\mathcal{S}_c)=0$ indicates that $e\in Conv(\mathcal{S}_c)$ and $I(e|\mathcal{S}_c)=1$ otherwise.

We then project $e$ onto the line connecting the two closest points $(s_1, s_2) \in \mathcal{S}_c$. We retrieve $(s_1, s_2)$ by computing the Euclidean distance for the regressed encodings with all elements of $\mathcal{S}_c$ and taking the two elements with the smallest distance. The error then is equal to the length of the vector rejection 
\begin{equation}
    \pi(e | s_1, s_2) = (s_1 - e) -  \frac{(s_1 - e)^T (s_2 - e)} {||s_2 - e||^2_2}(s_2 - e).
\end{equation}
\begin{figure}[t!]
    \centering
    \includegraphics[width=\linewidth]{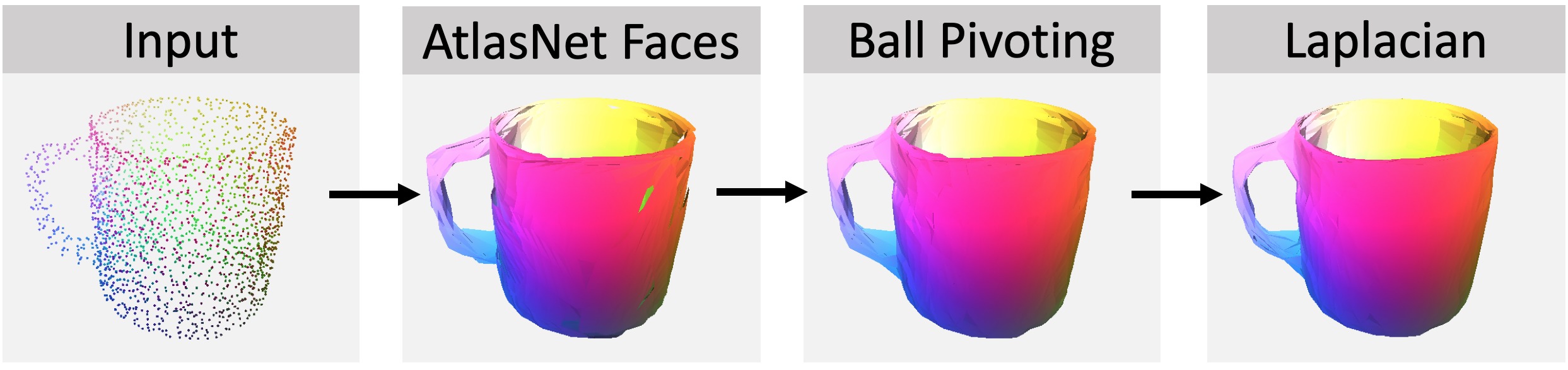}
    \caption{\textbf{3D Point Cloud Meshing.} To mesh our prediction, we make use of the connectivity implied by AtlasNet and fill remaining holes with the ball-pivoting algorithm. Finally, we smooth the result using the Laplacian filter.}
    \label{fig:meshing}
\end{figure}
\noindent  The final loss for one sample $e$ then amounts to
\begin{equation}
\scalemath{1}{
    \mathcal{L}_{reg}(e | \mathcal{S}_c, s_1, s_2) = I(e | \mathcal{S}_c) \cdot ||\pi(e | s_1,s_2)||_2.
}
\end{equation}
\smallskip

\subsubsection{Meshing of 3D Point Clouds} 
After estimating the shape as a point cloud, we can optionally also compute the associated mesh, \ie~the triangles of the model. Since AtlasNet samples 3D points uniformly from primitives, the triangles for each primitive can be easily inferred from the sampling. This facilitates meshing and allows a natural incorporation into our loss formulation, since we directly operate on point clouds. Unfortunately, the output mesh often exhibits holes. In order to fill these, we employ the ball-pivoting algorithm \citep{bernardini1999ball} and simply merge the output triangles. Finally, to reduce noise, we run one iteration of the Laplacian smoothing filter. The overall meshing process is also visualized in Fig.~\ref{fig:meshing}.

\subsection{3D Point Cloud Alignment}
Recent works \citep{Manhardt2019,Xiang2018,Simonelli2019} have shown that directly optimizing for the desired final target generally leads to superior results compared to enforcing separate loss terms for each regression target.
Motivated by this, we propose a novel loss directly aligning our regressed 3D shape using the predicted 6D pose with the scene.
\begin{figure}[t]
    \centering
    \includegraphics[width=\linewidth]{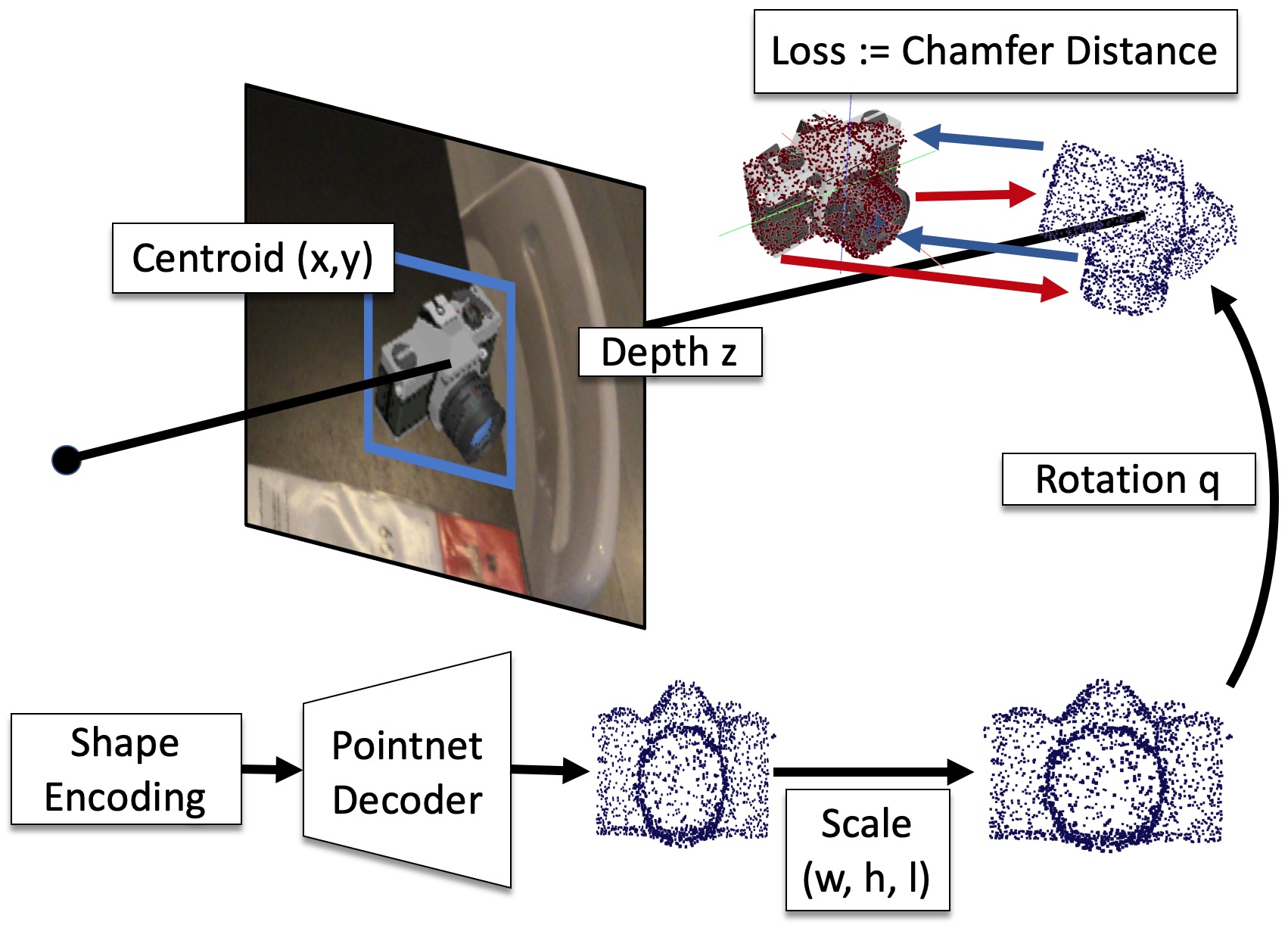}
    \caption{\textbf{3D Point Cloud Loss.} Given the outputs from our network, we first retrieve the detected object's shape from the AtlasNet decoder. We then scale it to absolute size before transforming it into the scene with the predicted rotation and translation. We employ the Chamfer distance between the ground truth point cloud and our prediction to enforce an optimal alignment in 3D.}
    \label{fig:pcd_loss}
\end{figure}
Given the egocentric 3D rotation as 4D quaternion $q$ and 3D translation $t = K^{-1} z\left( x, y, 1\right)^T$, together with the shape encoding $e$, the decoder $\mathcal{D}_c$, and the scale $( w, h, l)$, we compute the shape of the detected object and transform it to the 3D camera space to obtain the point cloud
\begin{equation}
    p_{3D} \coloneqq q \cdot \left[ \left( \begin{array}{ccc} w \\ h \\ l \end{array} \right) \cdot \mathcal{D}_c(e) \right] \cdot q^{-1} + K^{-1} \left( \begin{array}{ccc} x \cdot z \\ y \cdot z \\z \end{array}\right),
\end{equation}
with $K$ being the camera intrinsic matrix. We then measure the alignment against the ground truth point cloud $\bar{p}_{3D}$ using the Chamfer distance with
\begin{equation}
    \bar{p}_{3D} \coloneqq \bar{R}\bar{p} + \bar{t}.
\end{equation} Thereby, $\bar{R}$ and $\bar{t}$ denote the ground truth 3D rotation and translation and $\bar{p}$ denotes the ground truth point cloud computed by uniformly sampling 2048 points from the CAD model. The loss for 3D alignment is calculated as
\begin{equation}
\begin{aligned}
    \mathcal{L}_{3D} \coloneqq & \frac{1}{|p_{3D}|}\sum_{v \in p_{3D}}\min_{\bar{v} \in \bar{p}_{3D}}||v - \bar{v}||_{2} + \\ & \frac{1}{|\bar{p}_{3D}|}\sum_{\bar{v} \in \bar{p}_{3D}}\min_{v \in p_{3D}}||v - \bar{v}||_{2}.
\end{aligned}
\end{equation}
We also disentangle $\mathcal{L}_{3D}$ for our predictions, similar to \citep{Simonelli2019}. Therefore, we individually compute our 3D point cloud loss for each pose parameter (\ie~$q$, $c$, $z$, $(w,h,l)$, $e$), while taking the ground truth for the remaining parameters. The final 3D loss is then calculated as the mean over all individual loss contributions.

The overall loss is the sum of the loss for 3D alignment and shape regularization together with the loss for the object mask
\begin{equation}
    \mathcal{L}_{super} \coloneqq \mathcal{L}_{3D} + \mathcal{L}_{reg} + \mathcal{L}_{bce}.
\end{equation}
For mask prediction, we simply employ binary cross-entropy loss $\mathcal{L}_{bce}$. However, since each RoI mostly contains foreground pixels, the classification problem is not well balanced. Thus, to properly deal with class-imbalance, we separately apply the cross-entropy loss to all foreground and background pixels and then sum up both contributions.
\subsection{Domain Adaptation via Self-supervised Learning}

\subsubsection{Self-supervision for Pose \& Shape}

\begin{figure*}[t]
    \centering
    \includegraphics[width=\linewidth]{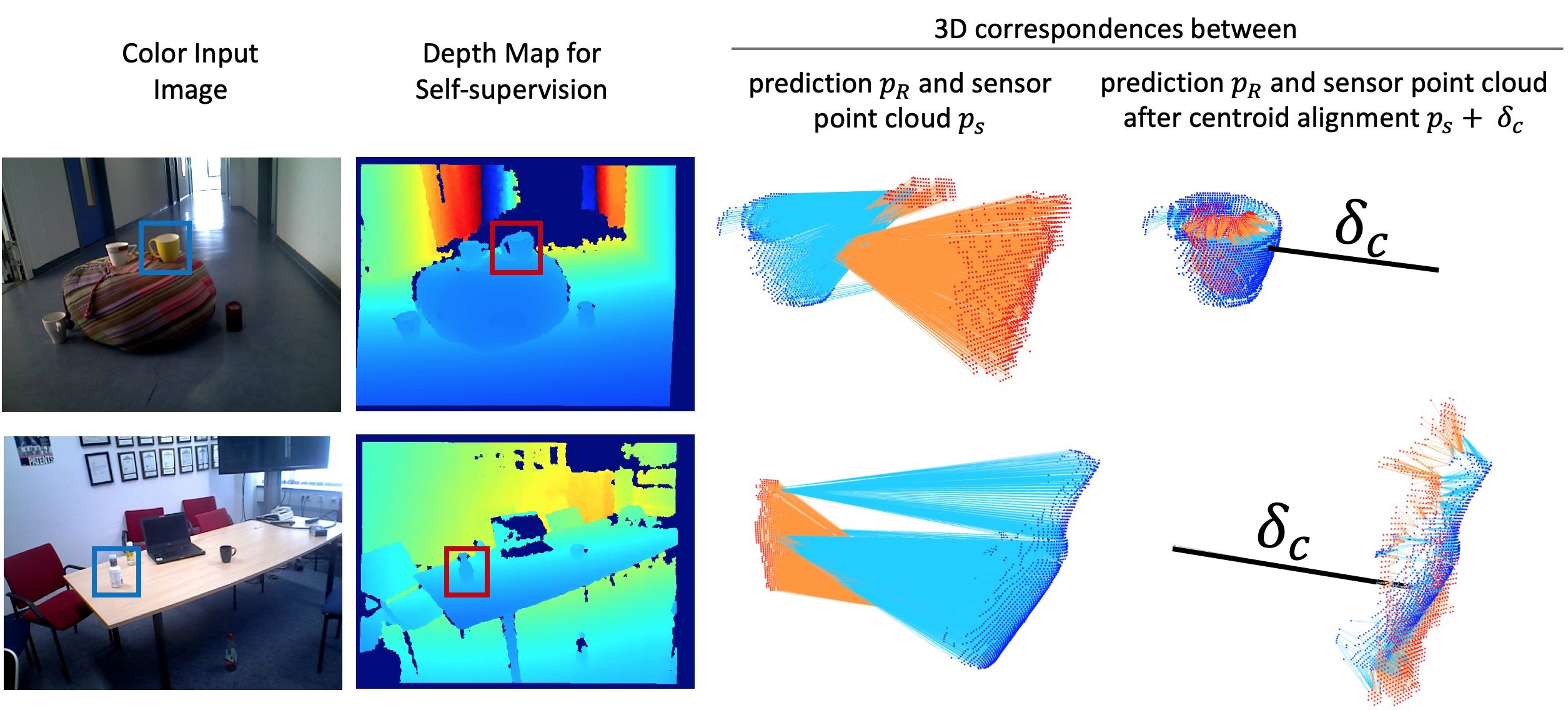}
    \caption{\textbf{3D Self-supervision.} Using the RGB image (1st column) we detect all objects in the scene and predict each object's 6D pose, metric shape and object mask $M_P$, from which we extract the visible point cloud $p_R$, shown in red. We also backproject the associated depth map (2nd column) \textit{w.r.t} to $M_P$ to retrieve the visible scene point cloud $p_S$, depicted in blue. Naively computing the Chamfer distance between $p_R$ and $p_S$ often converges in bad local minima due to weak correspondences as the shift in translation can be very large (3rd column). Hence, before we calculate the Chamfer distance, we instead first align the visible centroids according to $\delta_{c}$, providing more reliable correspondences and, thus, better supervision (4th column).}
    \label{fig:selfsupervision}
\end{figure*}

Since our proposed loss $\mathcal{L}_{super}$ for monocular class-level 6D pose estimation requires annotated data, which is difficult and time-consuming to collect, we train the network on synthetic samples only. Unfortunately, this leaves us with a domain gap towards the real world. To address this issue, \citet{wang2020self6d} recently proposed $\self$, in which they leverage real unlabeled RGB-D data to transfer the knowledge about the 6D pose from the synthetic to the real domain.

In this work, we adapt $\self$ to the problem of class-level 6D pose estimation to bridge the domain gap towards the real world. Provided the egocentric rotation $q$, the 3D translation $t$, and the 3D mesh $\mathcal{M}=(V,E)$ together with the camera matrix $K$, $\self$ renders a triplet consisting of RGB and depth image as well as the object mask in a differentiable manner. In order to fully rely on the raw sensor acquisitions without the need for 3D CAD models, we instead harness our predicted mesh vertices 
\begin{equation}
V = \left(\begin{array}{ccc} w \\ h \\ l \end{array} \right) \cdot \mathcal{D}_c(e)
\end{equation}
in metric scale and derive the triangles $E$ from the sampling of the 2D uv-map points in AtlasNet. Noteworthy, we always sample 2D locations on a uniform grid.

Since the improvement from the enforced loss on the RGB image was insignificant in \citep{wang2020self6d}, we dispense with the terms in the absence of sophisticated 3D color meshes. Therefore, we only render the pair of object masks and depth image
\begin{equation}
 \mathcal{R}(q, t, K, \mathcal{M}) = (D_R, M_R).
\end{equation}
For visual alignment, we thus only leverage the rendered mask $M_R$ in order to align the predictions with the scene $M_P$ according to \citep{jiang2019integral,wang2020self6d} with \begin{equation}\label{eg:loss_mask}
\begin{aligned}
\loss_{mask}  \coloneqq & -\frac{1}{|N_+|} \sum_{j\in N_+} M_{Pj} \log M_{Rj} 
- \\ & \frac{1}{|N_-|} \sum_{j\in N_-} \log (1 - M_{Rj}).
\end{aligned}
\end{equation} Thereby, $N_+$ and $N_-$ denote all foreground and background pixels with respect to $M_P$, respectively. 

Similar as in $\self$, we aim at establishing correspondences in 3D space in order to provide better supervision. \citet{wang2020self6d} aligns both visible point clouds after back-projection of the depth maps, $p_R = \pi^{-1}(D_R, M_R)$ and $p_S = \pi^{-1}(D_S, M_P)$, leveraging the Chamfer distance as objective function. Unfortunately, this does not work well in practice, since there is a high variance in translation (especially along the Z direction) due to the scale-distance ambiguity in monocular class-level pose estimation. As consequence, the 3D-3D correspondences do not represent anything meaningful for most predictions (as shown in Fig.~\ref{fig:selfsupervision}), making the training prone to convergence to bad local minima. Thus, prior to employing the Chamfer distance, we first coarsely align the visible 3D centroids of the point clouds   
\begin{equation}
c_R = \frac{1}{|p_R|} \sum_{v_R \in p_R} v_R \text{ and } c_S = \frac{1}{|p_S|} \sum_{v_S \in p_S } v_S
\end{equation}
\noindent according to $\delta_c = c_S - c_R$.

The final loss for geometrical alignment is then composed of the coarse alignment error $||\delta_c||_2$ and the Chamfer distance for fine alignment
\begin{equation}
\begin{aligned}
\loss_{geom} \coloneqq & \frac{1}{|p_S|}\sum_{v_S \in p_S}\min_{v_R \in p_R}\|v_S - v_R+ \delta_c \|_{2} +  \\ 
& \frac{1}{|p_R|}\sum_{v_R \in p_R}\min_{v_S \in p_S}\|v_S - v_R+ \delta_c\|_{2} + \\ & ||\delta_c||_2.
\end{aligned}
\end{equation}
\noindent Noteworthy, the presented loss for geometrical consistency is more biased towards punishing 3D translational errors. Hence, during self-supervision we build batches which contain both: annotated synthetic samples and real unlabeled samples. The overall self-supervision is, thus, a combination of the loss terms for visual and geometric alignment together with our original 3D point cloud alignment term
\begin{equation}
    \loss_{Self} \coloneqq \loss_{super} + \lambda_{mask} \loss_{mask} + \lambda_{geom} \loss_{geom}.
\end{equation}
\noindent Notice that in order to ensure that we only apply the loss for correct detections, we employ a very high detection threshold of 0.85 and only compute the loss on the remaining confident samples. Moreover, we also guarantee that at least 90\% of the masked pixels possess depth data and filter out all detections at a distance larger than 2.5 meters. We additionally remove outliers from the scene point cloud via region growing of the centroids on the depth map and statistical outlier removal on the point clouds. Despite the use of depth during self-supervision, the final inference is still fully monocular.
\smallskip

\subsubsection{Training Data for Self-supervision}
\begin{figure}[t!]
    \centering
    \includegraphics[width=0.49\linewidth]{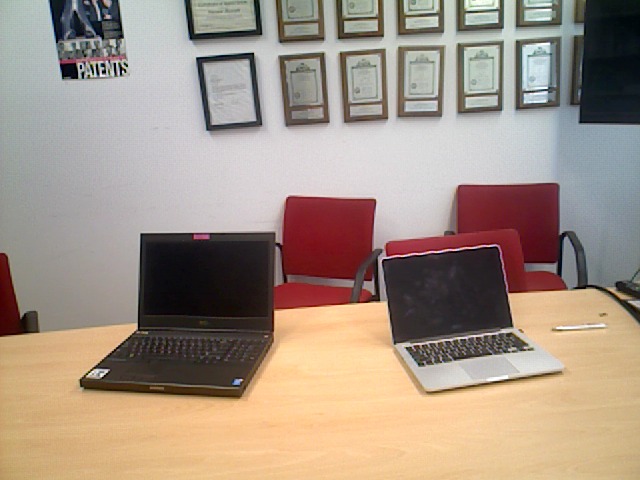}
    \includegraphics[width=0.49\linewidth]{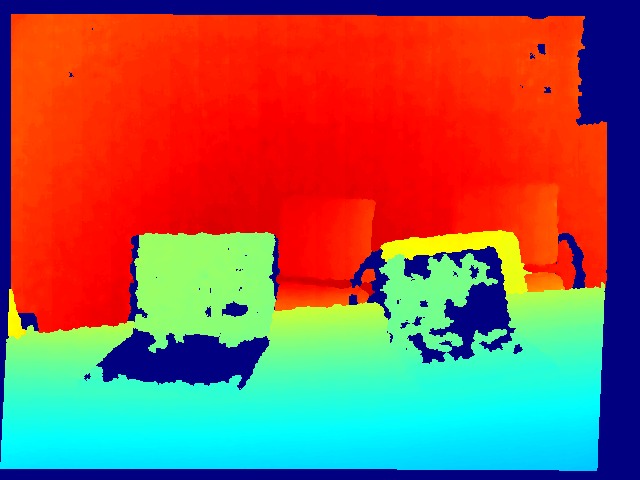} \\
    \includegraphics[width=0.49\linewidth]{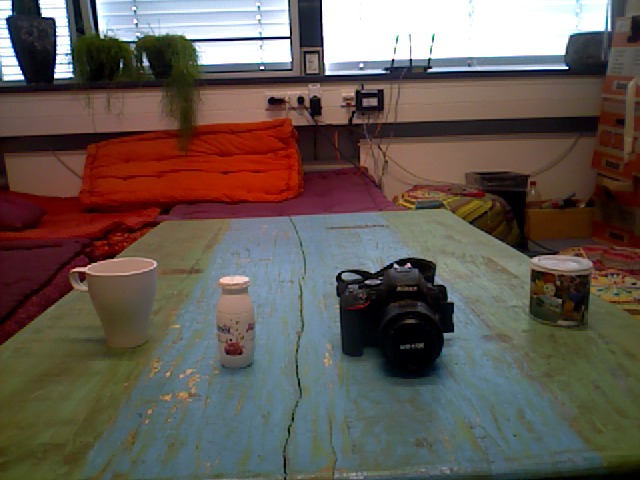}
    \includegraphics[width=0.49\linewidth]{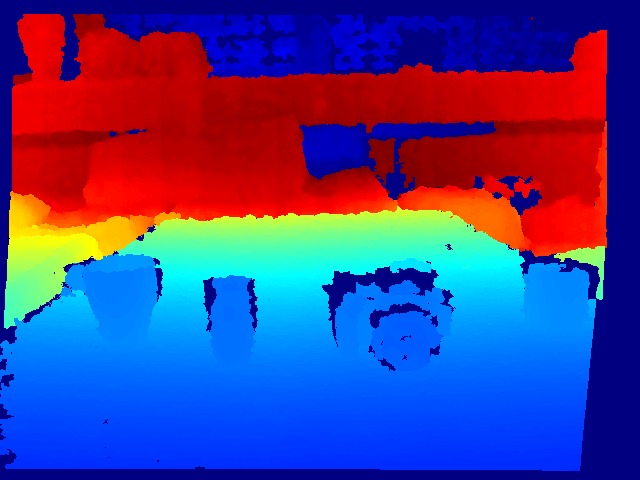} \\
    \includegraphics[width=0.49\linewidth]{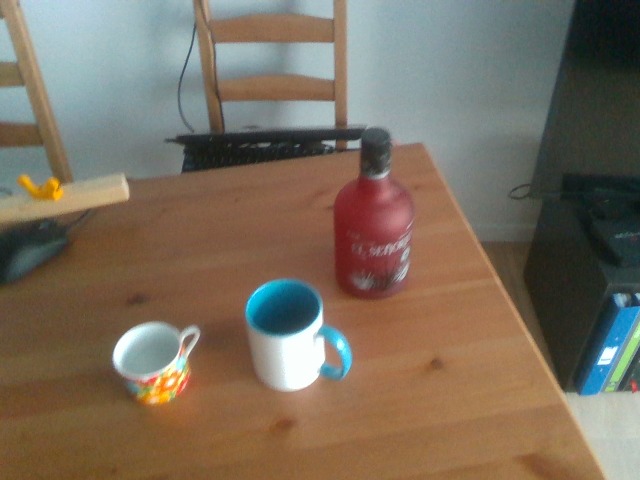}
    \includegraphics[width=0.49\linewidth]{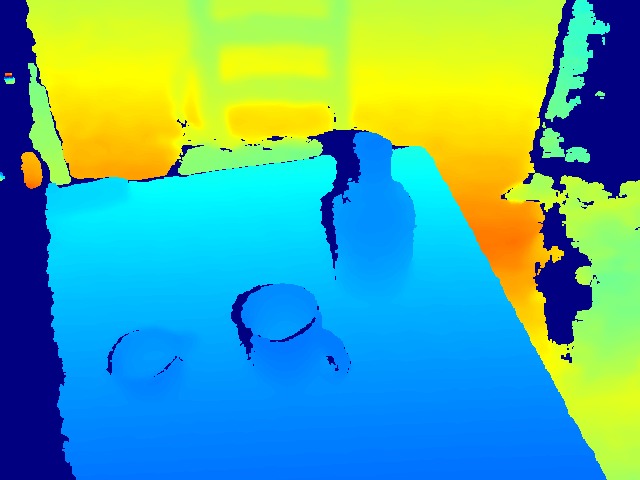}
    \caption{\textbf{Training data.} Exemplary samples of the recorded training RGB (left) and depth (right) images for self-supervision.}
    \label{fig:train_samples}
\end{figure}
In order to train with our self-supervised loss formulation, we recorded over 30k unlabeled RGB-D samples. We leveraged multiple different calibrated consumer RGB-D sensors based on structured light (\textit{e.g.} Primesene, Orbbec Astra) or stereo vision (\textit{e.g.} Intel RealSense). The data contains several different objects, each belonging to one of the six object classes from NOCS \citep{wang2019normalized}, \textit{i.e.} Bottle, Bowl, Can, Camera, Laptop, and Mug. Noteworthy, to make the proposed approach more applicable, we do not post-process (such as hole filling) the recorded data. A few exemplary samples can be found in Fig.~\ref{fig:train_samples}. The recorded data can be downloaded at \textcolor{blue}{\href{https://forms.gle/E89Asu3YDkL1WJEj6}{https://forms.gle/E89Asu3YDkL1W}}\-\textcolor{blue}{\href{https://forms.gle/E89Asu3YDkL1WJEj6}{JEj6}}, as we believe that self-supervision is an very important direction in the field of class-level 6D pose estimation.
\section{Evaluation}

In this section we first introduce our implementation details and demonstrate the evaluation protocol we followed. Afterwards, we present an ablation study on quality of the estimated shapes and the impact of each loss term during supervised and self-supervised learning and, finally, constitute our quantitative and qualitative results.

\subsection{Implementation Details}
We implemented our method in PyTorch \citep{paszke2019pytorch} and trained all models on a Nvidia Titan Xp GPU with a batchsize of 8 for 200k iterations using ADAM optimizer and a learning rate of 0.0001. We decay the learning rate after 20k, 130k, and 170k iterations by a factor of 0.1 each time. When training from scratch, directly applying the Chamfer distance turns out to be unstable due to potential convergence to local minima. Hence, we start with a warm-up training in which we compute the $L_1$-norm between each component and the ground truth using \citep{kendall2018multi} to weight the different terms.

Since annotating 6D pose, 3D scale, and 3D mesh is very difficult and time consuming, we decided to solely rely on synthetic data from NOCS for training. Nonetheless, to keep the domain gap small, we also sample with a probability of 35\% images from COCO \citep{coco_eccv14}, however, only back-propagate the 2D loss for these samples. Further, when we evaluate on the real test set, we additionally report each method after fine-tuning them for another 10k iterations on the real training data. Similarly, we also train with our self-supervision exactly for 10k iterations.

\subsection{Evaluation Protocol}

\subsubsection{NOCS Dataset}
For training and evaluation, we use the recently introduced NOCS dataset for class-level 6D pose estimation \citep{wang2019normalized}. It consists of about 270k synthetic training images and 25k synthetic validation images. \citep{wang2019normalized} employs a mixed-reality approach to render objects from ShapeNet \citep{chang2015shapenet} onto detected planes in real images. Additionally, they provide approximately 2.5k real test and 4.5k real training images. Overall, the dataset encompasses objects from 6 different classes, \textit{i.e.} Bottle, Bowl, Camera, Can, Mug and Laptop.

\subsubsection{6D Pose Metrics} 
Since we are the first to introduce the task of 6D pose estimation and metric shape retrieval, we want to propose a new metric that jointly measures the performance on both tasks. Thus, we extend two of the most common metrics for 6D pose known as \emph{Average Distance of Distinguishable Model Points} (ADD) and \emph{Average Distance of Indistinguishable Model Points} (ADI) \citep{Hodan2016,Hinterstoisser2011}. On one hand, ADD measures whether the average deviation $m$ of the transformed model points is less than $10\%$ of the object's diameter 
\begin{equation}
    m = \underset{x \in \mathcal{M}}{\avg} ||(Rx + t) - (\bar{R}x + \bar{t})||_2,
\end{equation}
where $\mathcal{M}$ denotes the set of points for the given CAD model. On the other hand, 
ADI extends ADD for symmetries, measuring error as the mean distance to the \emph{closest} model point
\begin{equation}
    m = \underset{x_{2} \in \mathcal{M}}{\avg}
    \min_{x_{1} \in \mathcal{M}} ||(Rx_{1} + t) - (\bar{R}x_{2} + \bar{t})||_2.
\end{equation} These metrics, however, are not applicable to our case since point sets for ground truth $\mathcal{\bar{M}}$ and predicted shape $\mathcal{M}$ differ and even possess differences in scale. To circumvent the need for direct correspondences and to be agnostic to scale discrepancies, we introduce the \emph{Average Distance of Predicted Point Sets} (APP) which extends ADI to be computed bidirectionally
\begin{equation}
    APP = 
    \begin{cases}
      1, & \text{if}\ m_{1} \leq \alpha \cdot d(\mathcal{M}) \land m_{2} \leq \alpha \cdot d(\mathcal{\bar{M}}) \\
      0, & \text{otherwise}
    \end{cases}
\end{equation}
where
\begin{equation}
 m_1 = \underset{x_{1} \in \mathcal{M}}{\avg}\min_{x_{2} \in \mathcal{\bar{M}}} ||(Rx_{1} + t) - (\bar{R}x_{2} + \bar{t})||_2 \\
\end{equation}
\begin{equation}
m_2 = \underset{x_{2} \in \mathcal{\bar{M}}}{\avg}
\min_{x_{1} \in \mathcal{M}} ||(Rx_{1} + t) - (\bar{R}x_{2} + \bar{t})||_2 
\end{equation}
and $d$ measuring the diameter of $\mathcal{M}$.  We employ $20\%$ and $50\%$ as thresholds for $\alpha$.

Since the related works do not incorporate shape prediction, we additionally compute the 3D IoU metric and $10\degree\&10cm$ metric to properly assess pose quality \citep{wang2019normalized}. For methods using depth, we additionally present the results for the more strict $5\degree\&5cm$ metric. Similar to previous works \citep{Simonelli2019,wang2019normalized}, we present all results computing the mean Average Precision (AP), measuring the area underneath the Precision-Recall curve.

\subsubsection{Comparison with Different Loss Functions from Related Works} 
Optimally weighting multiple different loss terms is kno\-wn to be complicated and a research topic on its own \citep{kendall2018multi,manhardt2018deep}. It is particularly challenging when the loss terms optimize two terms of different unit scales that cannot be easily compared such as rotation and translation. Consequently, we propose instead to directly measure the misalignment in 3D. To show the benefit of the proposed loss, we trained our network using only the $L_{1}$ loss for each prediction. To this end, we set each weighting component $\lambda=1$ (Uniform Weighting). We later employed a more elaborate training strategy which involves learning the different loss weights \citep{kendall2018multi} (Multi-Task Weighting).

We additionally implemented the loss function of the two most relevant works from autonomous driving for monocular 3D object detection on top of our network. In particular, we train our network to optimize the alignment of the bounding box corners (3D Bbox) \citep{Manhardt2019} and for the disentangled version from \citep{Simonelli2019} (Dis 3D Bbox). We chose these two works since, they estimate the full 6D pose instead of only predicting one angle for 3D rotation, as most related works do \citep{ding2020learning, chen2020monopair}.

\subsection{Ablation Study}

In order to evaluate how well our model estimates the shape of detected objects, we crop out the ground truth RoI from all images in the validation set and compute the Chamfer distance from each predicted shape as well as the mean $m_c$ (see Eq.~\ref{mean_shape}) to the corresponding ground truth. Notice that the shapes in the validation split of the NOCS dataset have not been seen before during training. Table~\ref{tab:shape_comparison} shows the mean Chamfer distance for each object class, respectively. The distance from predicted shape to ground truth is consistently lower, indicating that the model does indeed predict meaningful shapes for a specific image and object instance rather than trivially giving the mean. 
Qualitative examples of a bottle and a cup are given in the figures accompanying Table \ref{tab:shape_comparison}. The predicted bottle clearly exhibits the distinct corners and edges seen in the ground truth. In contrast, the mean bottle shows a very rounded out surface. Further, the predicted cup on the right shows the bowl-shape of the ground truth while the mean is more cylindrical.

\begin{table}[t]
    \centering
    \includegraphics[width=0.49\linewidth]{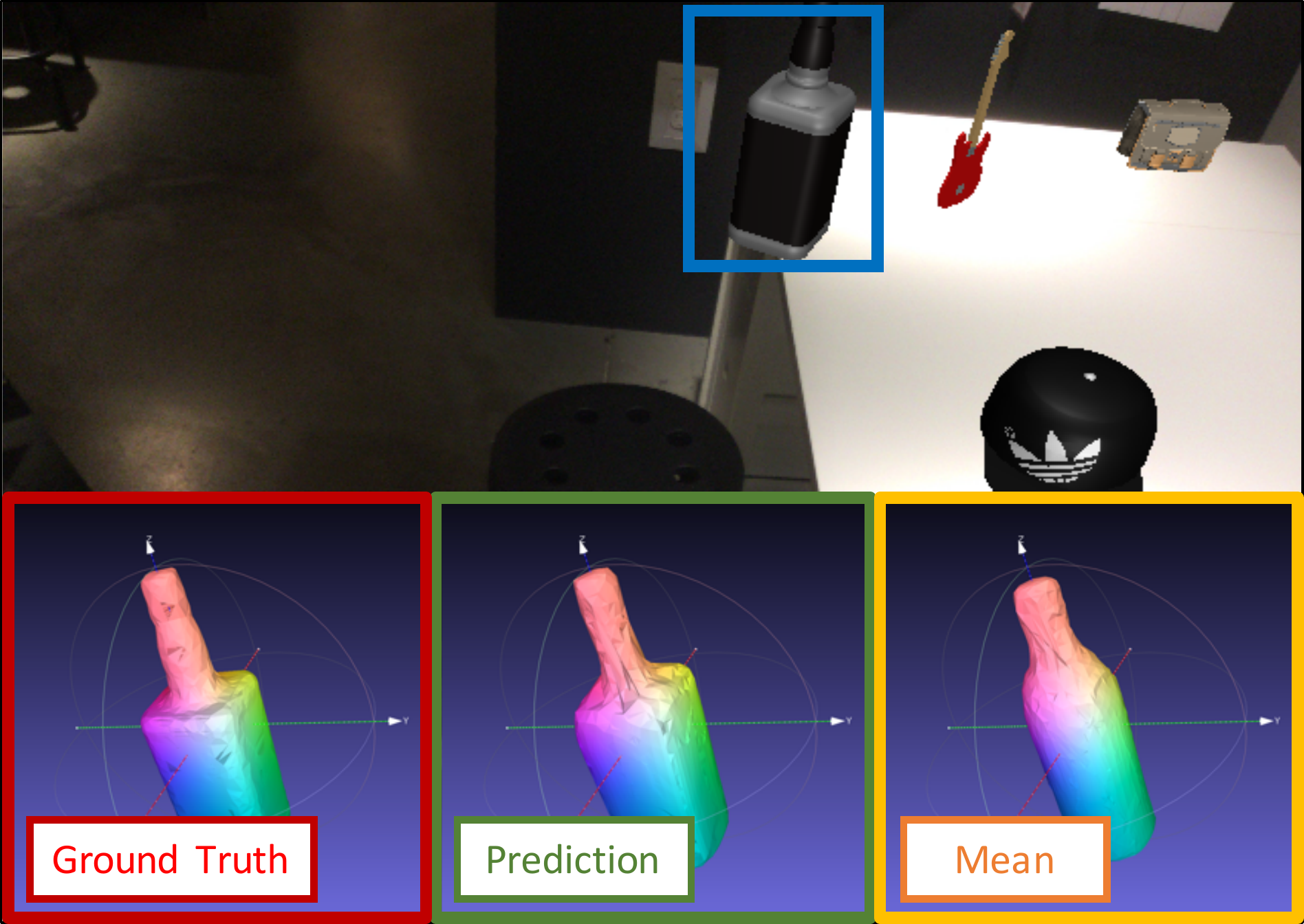}
    \includegraphics[width=0.49\linewidth]{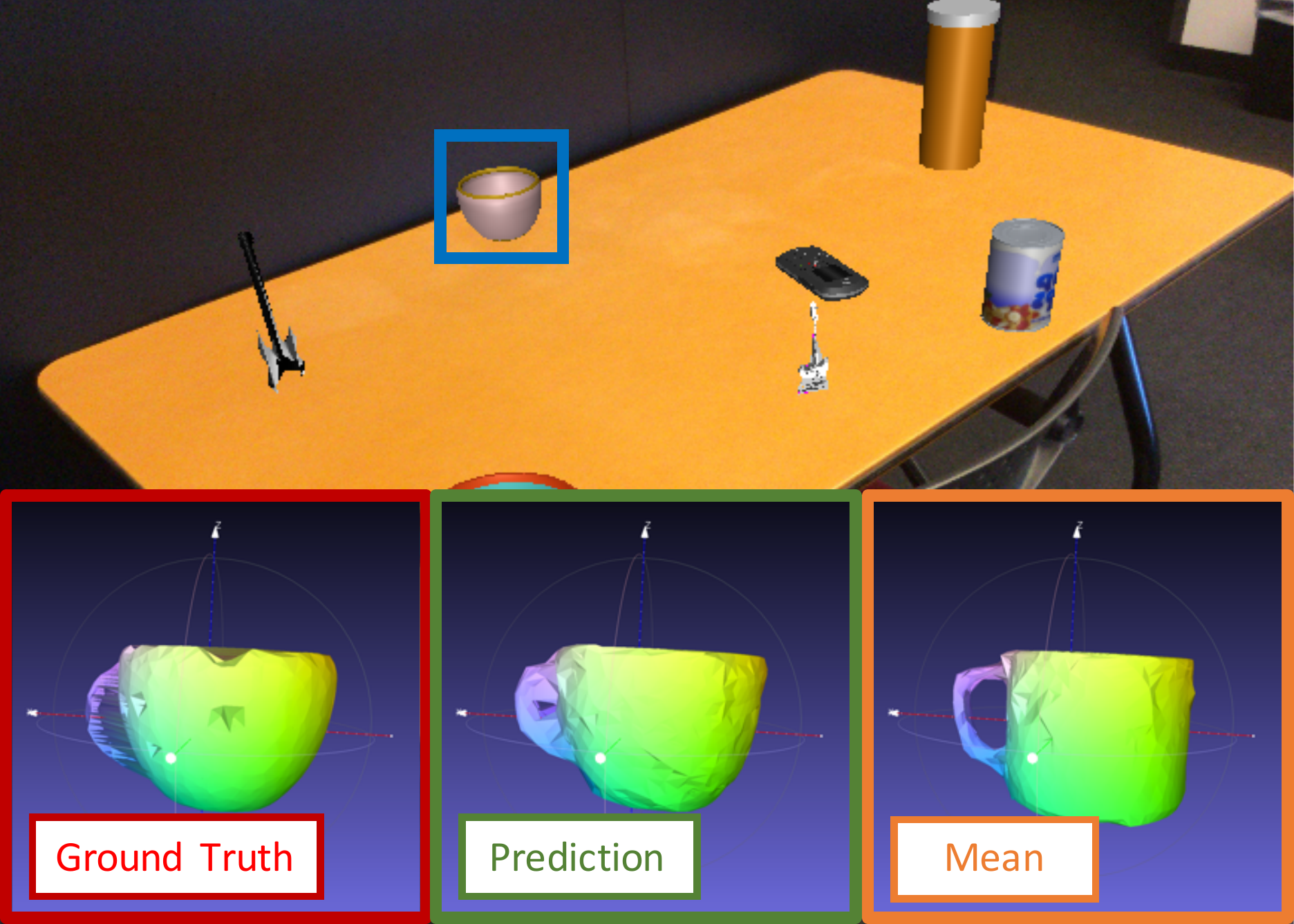}\\
    \scalebox{1.02}{
        \setlength{\tabcolsep}{1.5em}
        \begin{tabular}{|c||c|c|}
        \bottomrule
        \multirow{2}*{Object}  & \multicolumn{2}{c|}{Mean Chamfer Distance in mm $\downarrow$} \\
        & Mean Shape & Predicted Shape  \\
        \Xhline{1pt}
        Bottle  & 0.1028 & \textbf{0.0815} \\
        Bowl & 0.0892 & \textbf{0.0536} \\
        Can & 0.0161 & \textbf{0.0139} \\
        Camera & 0.1304 & \textbf{0.1026} \\
        Cup & 0.0481 & \textbf{0.0351} \\
        Laptop  & 0.3579 & \textbf{0.3117} \\
        \toprule
        \end{tabular}
    }
    \scalebox{0.92}{
        \setlength{\tabcolsep}{1.5em}
        \begin{tabular}{|c||c|c|}
            \bottomrule
            \multirow{2}*{$\cps$ w/ ICP}& 3D IOU  @ & 5\degree \& 5cm/ \\
            &  (0.25 / 0.5) $\uparrow$ & 10\degree \& 10cm $\uparrow$ \\
            \Xhline{1pt}
            Mean Shape & 89.8 / 67.4 & 37.6 / 57.7 \\
            Predicted Shape & \textbf{90.2} / \textbf{70.4} & \textbf{42.8} / \textbf{63.8} \\
            \toprule
        \end{tabular}
    }
    \caption{Comparison of predicted and mean shapes to the associated ground truth shapes. For each class, the mean Chamfer distance of the shapes estimated by our model is lower than that of the mean over all shapes. When estimating pose, leveraging our estimated models for ICP leads to better results.}
    \label{tab:shape_comparison}
\end{table}

Following standard practice, we also evaluate class-level 6D pose estimation after refining the poses with ICP on the accompanying depth data. Thereby, we compare ICP leveraging the predicted shapes with ICP employing the mean shapes. While for rather simple objects the difference is negligible, for more complex shapes the discrepancy is vast. For more strict metrics, such as IoU@0.5 and $5\degree\&5cm$, the relative performance drops around 3\% and 5\%. This clearly shows that our shapes are more accurate than the corresponding mean shape for each class and can be leveraged for more reliable 6D pose estimation.

In Table~\ref{tab:pose_ablation} and~\ref{tab:pose_ablation_self}, we want to demonstrate the individual loss contributions. Therefore, we train and evaluate $\cps$ on the synthetic training and validation dataset and the proposed self-supervision $\cps$++ on the recor\-ded unlabeled real training data and the real test data, always turning off one loss component during training. 

\begin{table*}[t]
    \centering
    \scalebox{1}{
        \begin{tabular}{|c||c|c|c|}
        \bottomrule
        Method & 3D IOU @ (0.25 / 0.5) & 10\degree \& 10cm & 3D APP @ (0.2 / 0.5) \\
             \Xhline{1pt}
            $\cps$ w/o disentangling  & 21.4 / 5.1 & 17.4 & 14.0 / 42.1 \\ 
            $\cps$ w/o $\mathcal{L}_{reg}$ & 21.9 / 5.4 & 26.5 & 13.0 / 41.7 \\
           $\cps$ & \textbf{29.0} / \textbf{8.7} & \textbf{31.7} & \textbf{19.1} / \textbf{49.6} \\
            $\cps$++ & 26.7 / 8.1 & 27.4 & 17.8 / 45.2 \\
            \toprule
        \end{tabular}
    }
    \caption{Ablation study on the synthetic validation dataset from \citep{wang2019normalized}. We report AP scores for 3D IoU, rotation and translation as well as APP.}
    \label{tab:pose_ablation}
\end{table*}

The ablation on the synthetic data shows that punishing encodings out of their associated spaces makes training more robust and leads to better results. Especially, complicated shapes such as \emph{Camera} (7.2\% \textit{vs.} 3.4\% for $10\degree\&10cm$ metric) mostly benefit from the regularization term. We report individual results for each object in the supplementary material. Moreover, disentangling the individual loss components further stabilizes optimization, leading to a significant increase in performance across all metrics. Interestingly, when evaluating our method after self-supervision on the synthetic data, the performance significantly degrades. Sin\-ce $\cps$ is purely trained on the synthetic domain, the network is performing particularly well on those samples. After self-supervised learning the network generalizes better to real data, however, for the cost of losing performance on the synthetic domain.

\begin{table*}[t]
    \centering
    \scalebox{1.}{
        \begin{tabular}{|c|c|c|c||c|c|c|}
        \bottomrule
        & \multirow{2}*{Mask Loss} & \multicolumn{2}{c||}{Geometry Loss} & 3D IOU @ (0.25 / 0.5) & 10\degree \& 10cm & 3D APP @ (0.2 / 0.5) \\
        & & Chamfer & Centroid & & & \\
            \Xhline{1pt}
             $\cps$ & \multicolumn{3}{c||}{} & 43.7 / 14.0 & 16.5 & 30.8 / 64.0 \\
            \Xhline{1pt}
            \multirow{4}*{$\cps$++} & & \checkmark & \checkmark & 32.3 / 2.2 & 1.9 & 15.4 / 65.2 \\
             &  \checkmark& & & 31.3 / 9.5 & 4.9 & 22.2 / 53.6 \\
             & \checkmark & \checkmark & & 47.1 / 11.9 & 17.4 & 32.4 / 68.2 \\
             & \checkmark & \checkmark & \checkmark & \textbf{54.3} / \textbf{17.7} & \textbf{22.3} & \textbf{41.0} / \textbf{73.6} \\
            \toprule
        \end{tabular}
    }
    \caption{Ablation study on different loss terms for self-supervision on the real test dataset from \citep{wang2019normalized}. We report AP scores for 3D IoU, rotation and translation as well as APP.}
    \label{tab:pose_ablation_self}
\end{table*}

On the other hand, the results on the real data demonstrate that only the modified geometry loss is capable of improving the results and can thus successfully decrease the domain gap. In particular, without centroid loss the performance is similar to the performance before self-supervision. Noteworthy, when turning off either the mask loss or geometry loss, the results even fall behind the original performance of the purely synthetically trained model.

\subsection{Quantitative Evaluation}

\subsubsection{Synthetic Data Experiments}
Table~\ref{tab:pose_comparison_synthetic} shows evaluation results on the synthetic dataset published with \citep{wang2019normalized}. While all methods are at a similar level, $\cps$ still shows superior results with respect to most metrics, especially stricter ones such as IoU@0.5 and 10\degree \& 10cm. 

\begin{table*}[t]
    \centering
    \scalebox{1.}{
        \begin{tabular}{|c||c|c|c|}
        \bottomrule
        & 3D IOU @ (0.25 / 0.5) & 10\degree \& 10cm & 3D APP @ (0.2 / 0.5) \\
             \Xhline{1pt}
             Uniform Weighting & \textbf{29.2} / 7.5 & 26.9 & 17.0 / 49.2 \\
             Multi-Task Weighting \citep{kendall2018multi}  & 28.9 / 7.1 & 26.9 & 17.1 / \textbf{50.2} \\
             3D Bbox Loss \citep{Manhardt2019}   & 19.5 / 3.3 & 22.1 & -- / -- \\
             Dis 3D BBox Loss \citep{Simonelli2019}  & 28.4 / 6.6 & 17.7 & -- / -- \\ 
             \Xhline{1pt} 
            $\cps$ & 29.0 / \textbf{8.7} & \textbf{31.7} & \textbf{19.1} / 49.6 \\
            $\cps$++ & 26.7 / 8.1 & 27.4 & 17.8 / 45.2 \\
            \toprule
        \end{tabular}
    }
    \scalebox{1.}{
        \begin{tabular}{|c||c|c|c|}
        \bottomrule
        & 3D IOU @ (0.25 / 0.5) & 5\degree \& 5cm /10\degree \& 10cm & 3D APP @ (0.2 / 0.5) \\
             \Xhline{1pt}
            NOCS \citep{wang2019normalized} & \textbf{91.4} / \textbf{85.3} & 38.8 / 62.2 & -- / -- \\
            $\cps$ w/ ICP  & 90.2 / 70.4 & \textbf{42.8} / \textbf{63.8} & \textbf{89.0} / \textbf{91.3} \\
            $\cps$++ w/ ICP  & 89.4 / 63.4 & 33.6 / 49.6 & 88.2 / 91.2 \\
             \toprule
        \end{tabular}
    }
    \caption{State-of-the-art methods evaluated on the synthetic validation dataset from \citep{wang2019normalized}. We report AP scores for 3D IoU, rotation and translation as well as APP.}
    \label{tab:pose_comparison_synthetic}
\end{table*}

Surprisingly, the two baselines (on the top) that simply weight the individual loss terms achieve overall strong results, yet, $\cps$ still exceeds them. As for the AP score on 3D IoU and a threshold of 0.5, we outperform \citep{kendall2018multi} by $1.6\%$ and \emph{Uniform Weighting} by $1.2\%$ with a score of $8.7$. Similar results can be also observed for $10\degree\&10cm$ metric and APP. In particular, we outperform both by $4.8\%$ for $10\degree\&10cm$ and ca. $2\%$ for APP\@0.2 achieving a performance of $31.7\%$ and $19.1\%$, respectively.

Employing other recently proposed loss functions from \citep{Manhardt2019} and \citep{Simonelli2019} led to inferior results. For most metrics we can exceed their performance by more than $30\%$ of relative accuracy, proving that our point cloud alignment is unquestionably more effective than \eg aligning 3D bounding box corners. In contrast to autonomous driving, objects in robotics often exhibit ambiguities such as symmetries (\eg cans and bottles). Since \citep{Manhardt2019,Simonelli2019} are minimizing the misalignment of 3D bounding box corners, their methods are very sensitive to ambiguities.

To demonstrate the potential of $\cps$ for real robotic applications and for a fair comparison with the recently published RGB-D method NOCS \citep{wang2019normalized}, we also evaluate our method after refining the poses with ICP on the associated depth maps, on the basis of our predicted shapes. The corresponding results are depicted in the bottom table. When employing depth, our numbers increase significantly. In particular, our AP score with respect to APP and 3D IoU more than quadruples. In addition, for $10\degree\&10cm$ we can also double the results reported for monocular $\cps$. 
Moreover, when using RGB-D we reach state-of-the-art performance for class-level 6D pose estimation. While NOCS exceed us in terms of 3D IoU at a threshold of 0.5,
we can outperform them for the $5\degree\&5cm$ and $10\degree\&10cm$ metric with $42.8\%$ and $63.8\%$ in comparison to $38.8\%$ and $62.2\%$. Similar to before, despite the use of ICP, we experience a decrease in performance when evaluating $\cps$++ on the synthetic validation data due to worse initializations.

\subsubsection{Real Data Experiments}

We run evaluations on the real dataset of \citep{wang2019normalized}. The results are reported in Table~\ref{tab:pose_comparison_real}.

\begin{table*}[t!]
    \centering
    \includegraphics[width=0.24\linewidth,trim={0 0 1.6cm 1cm},clip]{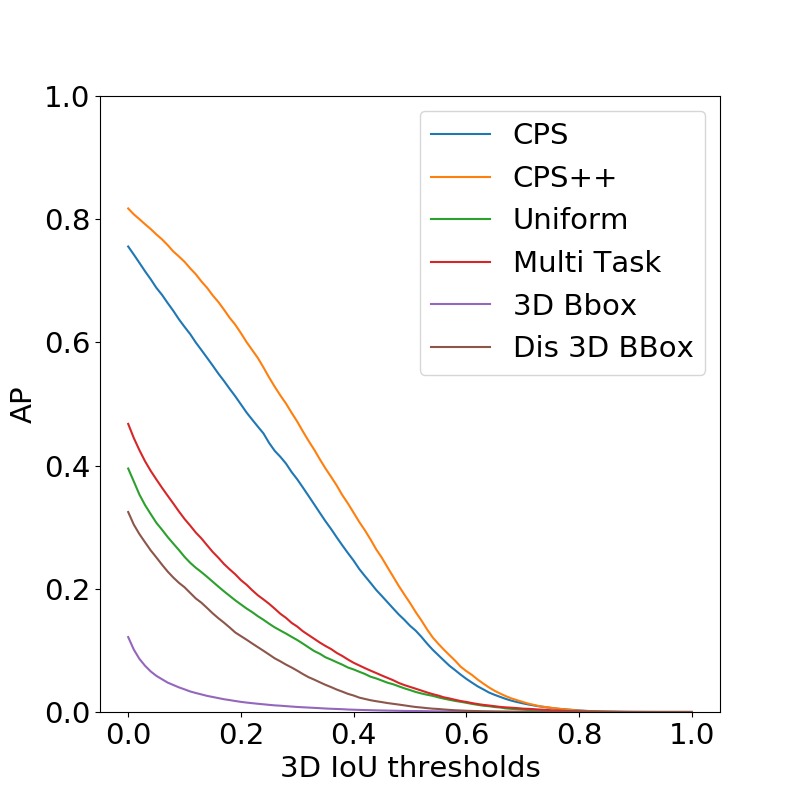}
    \includegraphics[width=0.24\linewidth,trim={0 0 1.6cm 1cm},clip]{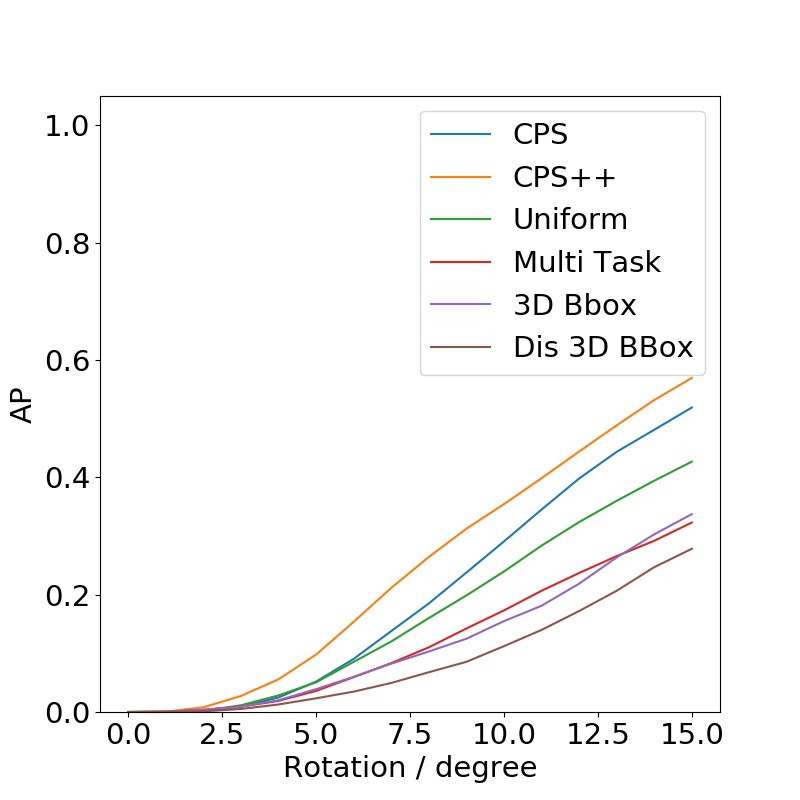}
    \includegraphics[width=0.24\linewidth,trim={0 0 1.6cm 1cm},clip]{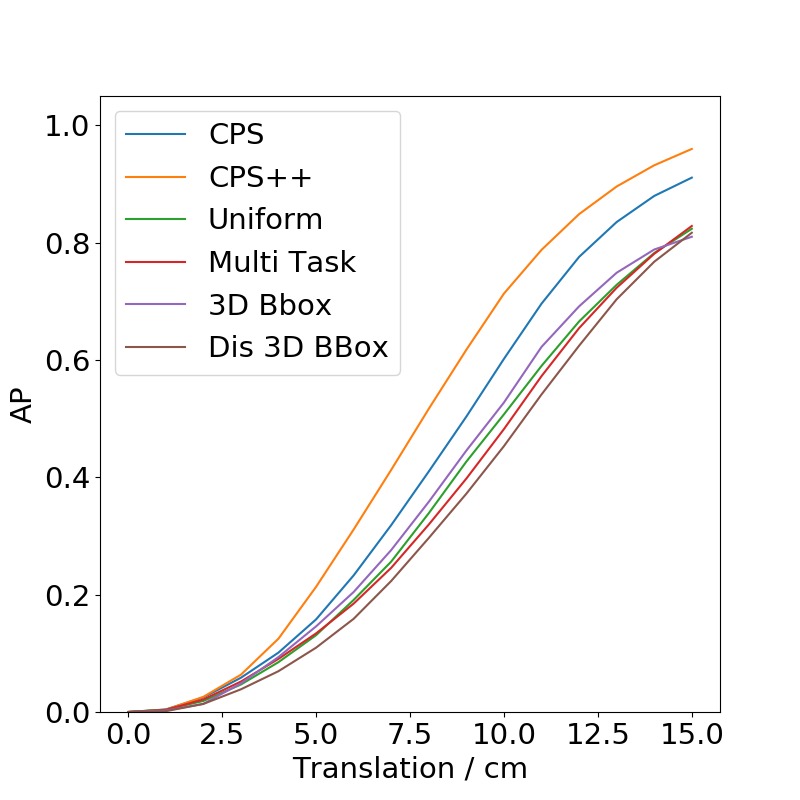} 
    \includegraphics[width=0.24\linewidth,trim={0 0 1.6cm 1cm},clip]{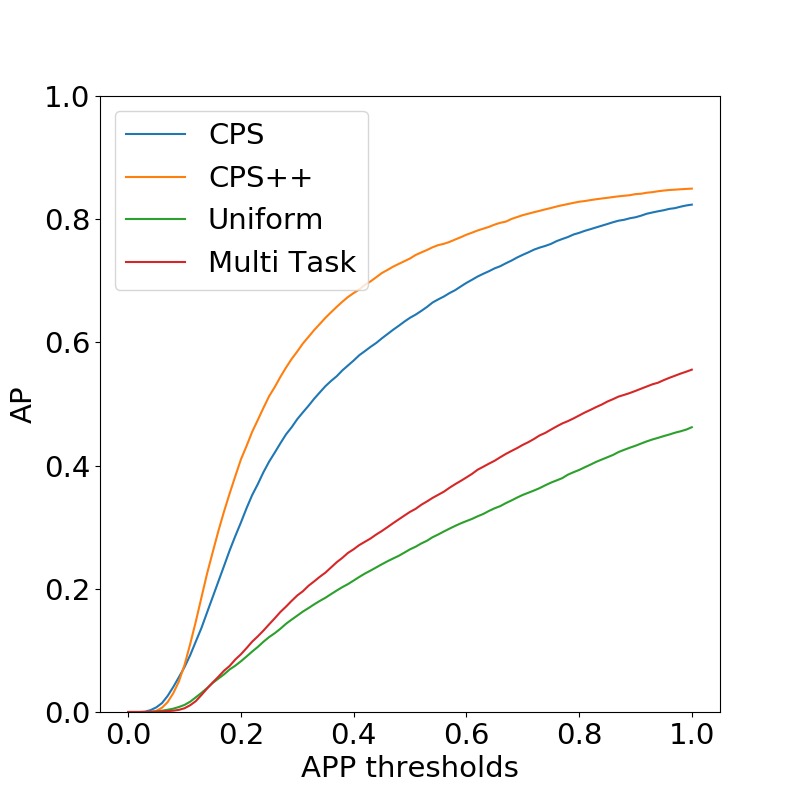}
     
    \includegraphics[width=0.24\linewidth,trim={0 0 1.6cm 1cm},clip]{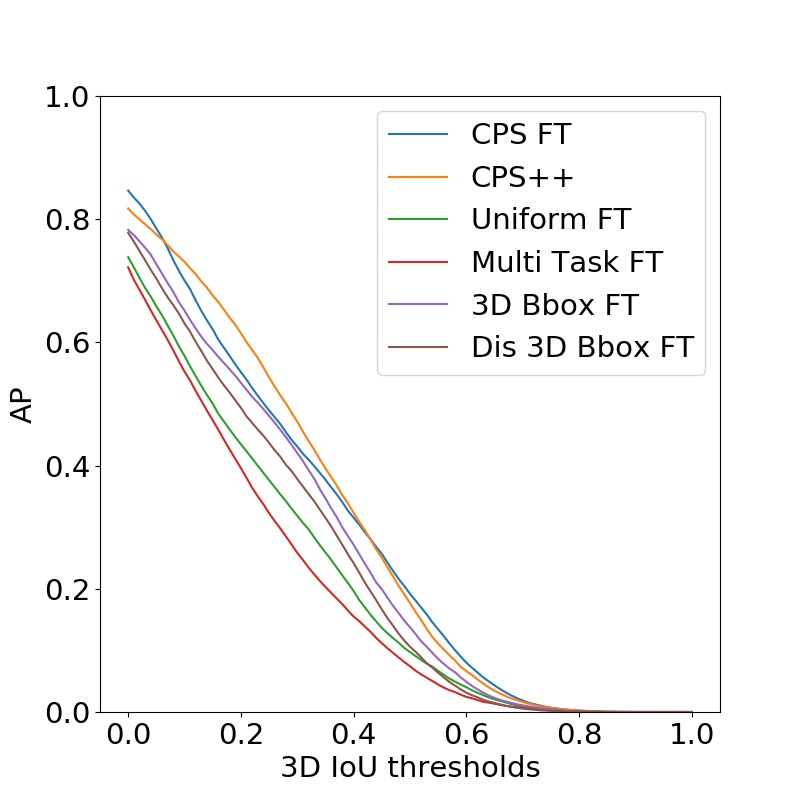}
    \includegraphics[width=0.24\linewidth,trim={0 0 1.6cm 1cm},clip]{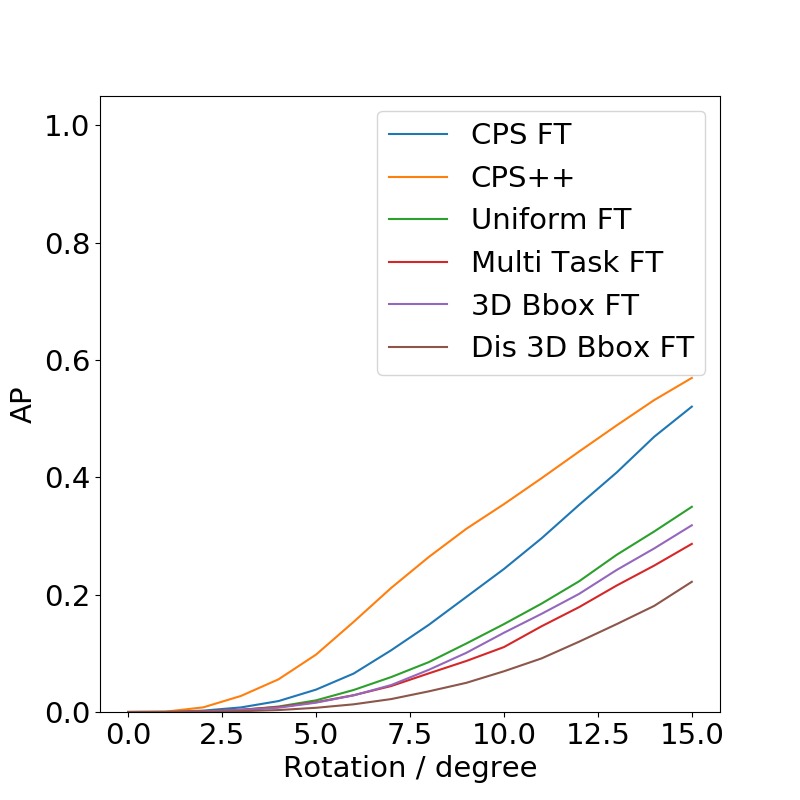}
    \includegraphics[width=0.24\linewidth,trim={0 0 1.6cm 1cm},clip]{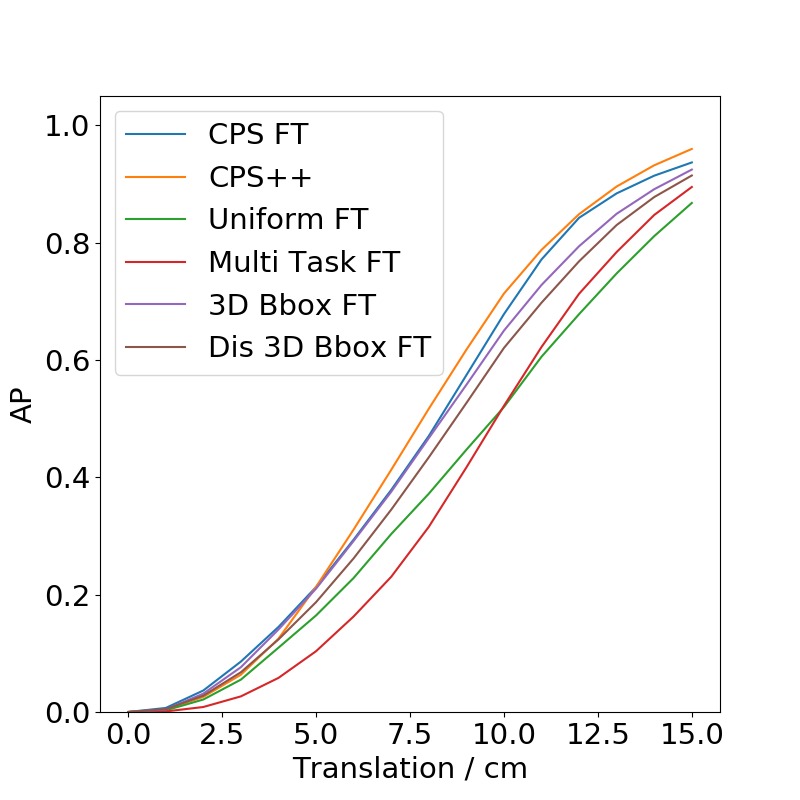}
    \includegraphics[width=0.24\linewidth,trim={0 0 1.6cm 1cm},clip]{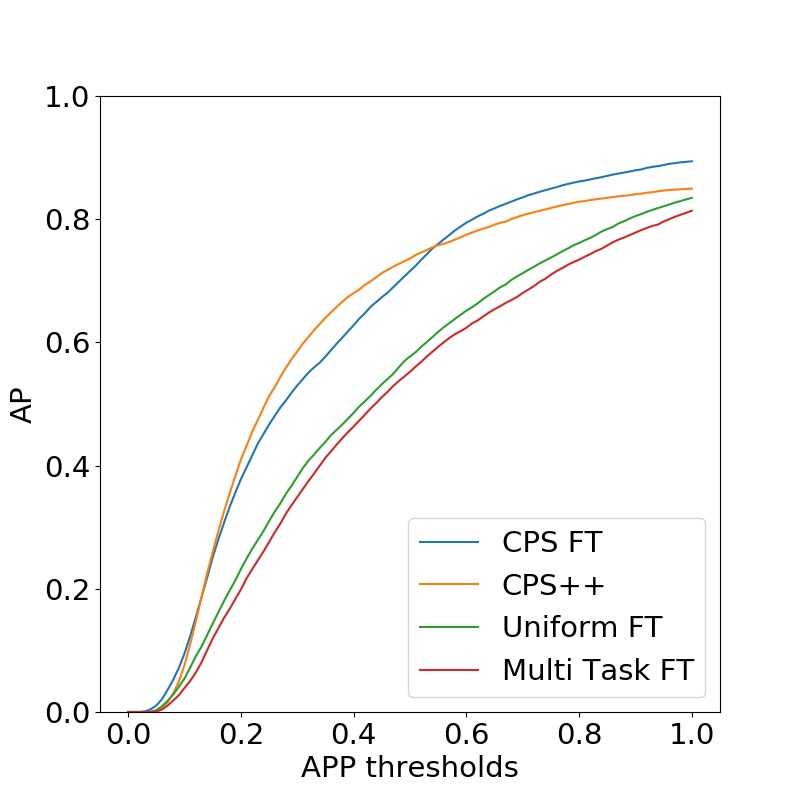}
    
    \scalebox{1.}{
        \setlength{\tabcolsep}{0.3em}
        \begin{tabular}{|c|c||c|c|c|}
        \bottomrule
        & FT Real & 3D IOU @ (0.25 / 0.5) &  10\degree \& 10cm & 3D APP @ (0.2 / 0.5) \\
             \Xhline{1pt}
             Uniform Weighting & & 14.3 / 3.6 & 15.0 & 8.2 / 26.4 \\
             Multi-Task Weighting \citep{kendall2018multi} & & 17.6 / 4.1 & 10.4 & 9.4 / 32.5 \\
             3D Bbox Loss \citep{Manhardt2019} &  & 1.2 / 0.2 & 11.2 & -- / -- \\
             Dis 3D Bbox Loss \citep{Simonelli2019} &  & 9.3 / 0.9 & 3.9 & -- / -- \\
             \Xhline{1pt}
             Uniform Weighting & \checkmark & 37.6 / 9.7 & 6.8 & 23.2 / 57.7 \\
             Multi-Task Weighting \citep{kendall2018multi}  & \checkmark  & 32.3 / 7.4 & 4.7 & 19.9/ 55.2 \\
             3D Bbox \citep{Manhardt2019}  & \checkmark   & 42.0 / 13.8 & 9.1 & -- / --\\
             Dis 3D BBox \citep{Simonelli2019}  & \checkmark  & 43.7 / 10.6 & 3.1 & -- / -- \\
             \Xhline{1pt}
             $\cps$ & &  43.7 / 14.0 & 16.5 & 30.8 / 64.0 \\
             $\cps$ & \checkmark &  48.9 / \textbf{19.2} & 14.7 & 37.8 / 71.6 \\
             $\cps$++ & & \textbf{54.3} / 17.7 & \textbf{22.3} & \textbf{41.0} / \textbf{73.6} \\
             \toprule
        \end{tabular}
    }
    \\
    \scalebox{1.}{
     \setlength{\tabcolsep}{0.3em}
        \begin{tabular}{|c|c||c|c|c|}
        \bottomrule
        & Real Data w Labels & 3D IOU @ (0.25 / 0.5) &  5\degree \& 5cm  / 10\degree \& 10cm & 3D APP @ (0.2 / 0.5) \\
        \Xhline{1pt}
             NOCS \citep{wang2019normalized} & & 57.6 / 41.0 & 3.3 / 17.1 & -- / -- \\
             NOCS \citep{wang2019normalized} & \checkmark & \textbf{84.9} /  \textbf{80.9} & 9.5 / 26.7 & -- / -- \\
             CASS \citep{chen2020learning} & \checkmark &  84.2 /  77.7 & 13.0 / 37.9 & -- / -- \\ 
             \Xhline{1pt}
             $\cps$ w/ ICP  & &  84.5 / 72.6 & \textbf{25.8} / 55.4 & \textbf{83.3} / \textbf{86.3} \\
             $\cps$++ w/ ICP & & 84.6 / 72.8 & 25.2 / \textbf{58.6} & 81.1 / 85.7 \\
             \toprule
        \end{tabular}
    }
    \caption[]{State-of-the-art methods evaluated on the real test dataset from \citep{wang2019normalized} Top: We plot AP scores for 3D IoU, rotation and translation, and APP with respect to increasing thresholds. Bottom: We report AP scores for 3D IoU, rotation and translation, and APP at commonly employed thresholds\footnotemark[1]. Notice that all methods leverage COCO to decrease the domain gap.
    }
    \label{tab:pose_comparison_real}
\end{table*}

While the performance difference on the synthetic data is rather small, for the real test dataset, the gap is very large. In particular, $\cps$ more than doubles all other methods for 3D IoU and APP. Also for $10\degree\&10cm$ $\cps$ comes out as superior. This indicates that $\cps$ is much stronger at generalizing than all other methods. For the synthetic training only case, we are even on par with NOCS without leveraging any depth information. 

When finetuning each network by another 10k iterations on the real labeled data, in order to address the domain gap, the related works are capable of almost closing the gap. Nonetheless, whereas 3D IoU is almost on par with our method, we clearly outperform them on $10\degree\&10cm$. When investigating the accompanying plots for each metric, we can deduce that mostly 3D translation improved from the finetuning. In contrast, the 3D rotation AP even slightly decreased. This can be attributed to the limited variation in rotation of the real test data, since it is a very large space and the number of samples are limited. Noteworthy, as \citep{wang2019normalized} computes 3D IoU with respect to the main axes, the drop in rotation accuracy is not strongly reflected there.

Further, while finetuning $\cps$ on real data leads again to small improvements, we can enhance the performance even further when instead leveraging our proposed self-supervision. For 3D IoU we can improve by $10.6\%$ and $3.7\%$ with $54.3\%$ and $17.7\%$ at a threshold of 0.25 and 0.5, respectively. Regarding $10\degree\&10cm$, we can report an AP of $22.3\%$ in comparison to $16.5\%$. Similarly, with respect to APP at 0.2 and 0.5, we can increase the AP by ca. $10\%$. These observations are also reflected by the accompanying graphs, as $\cps$++ exceeds all other monocular methods with respect to all metrics at any threshold.

\begin{figure*}[t]
    \centering
    \begin{tabular}{cccc}
        3D Bounding Box &  XZ-plane Projection &  XY-plane Projection& Recovered Mesh \\ 
        \includegraphics[width=0.22\textwidth]{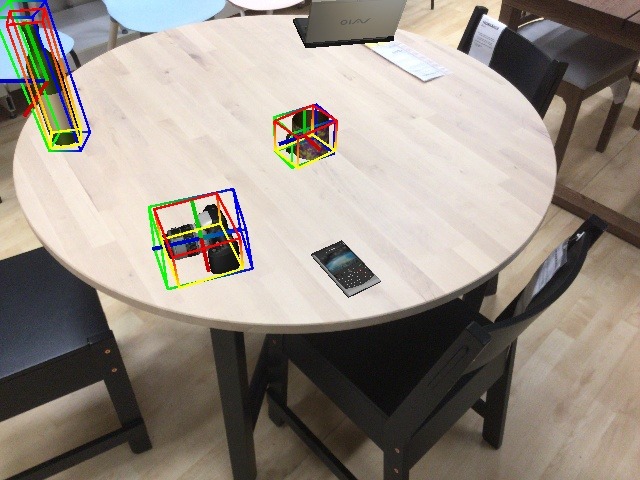} &
        \includegraphics[width=0.22\textwidth]{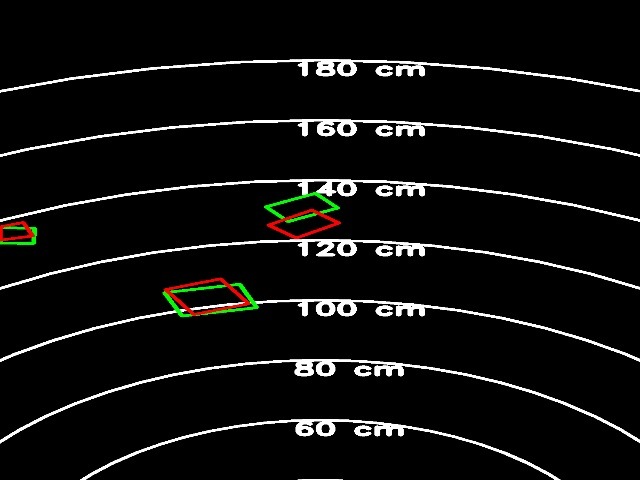} &
        \includegraphics[width=0.22\textwidth]{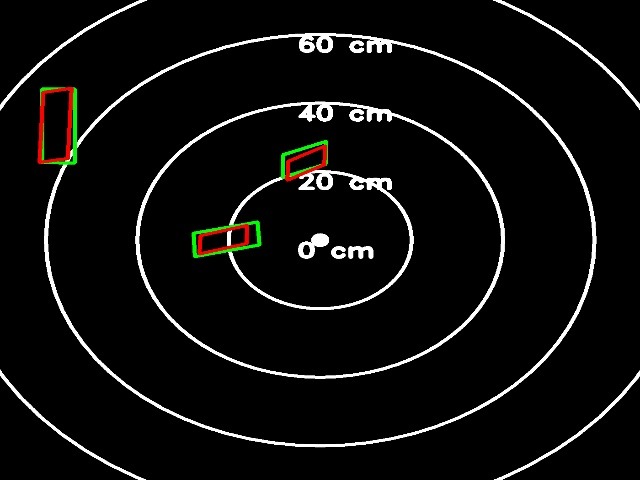} &  
        \includegraphics[height=82pt, width=84pt]{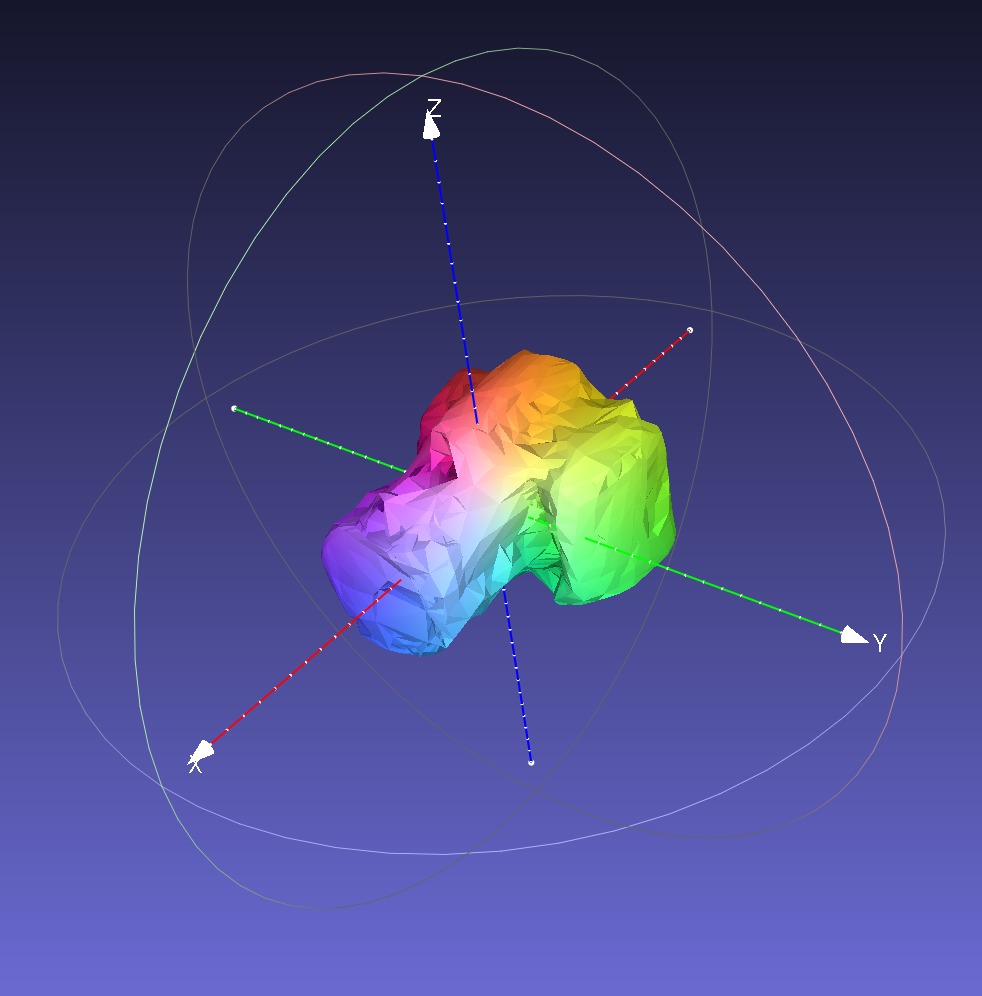} \\
        \includegraphics[width=0.22\textwidth]{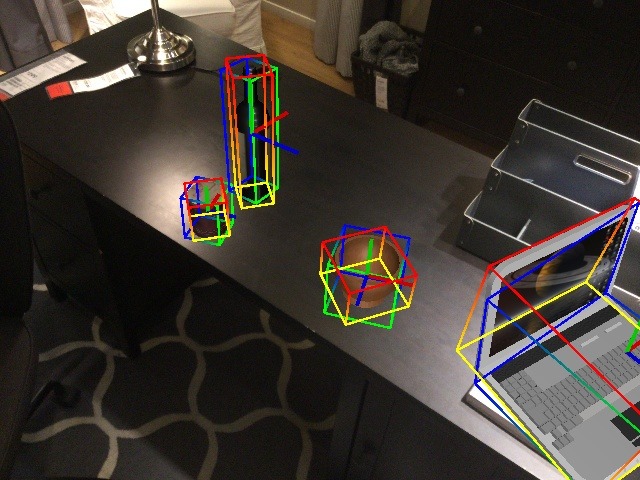} &
        \includegraphics[width=0.22\textwidth]{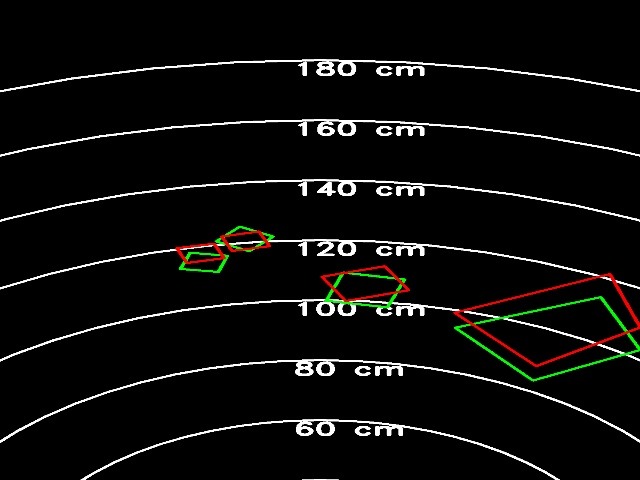} &
        \includegraphics[width=0.22\textwidth]{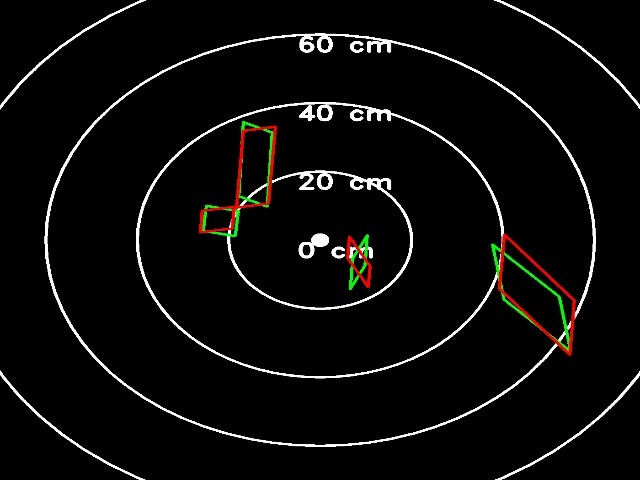} &  
        \includegraphics[height=82pt, width=84pt]{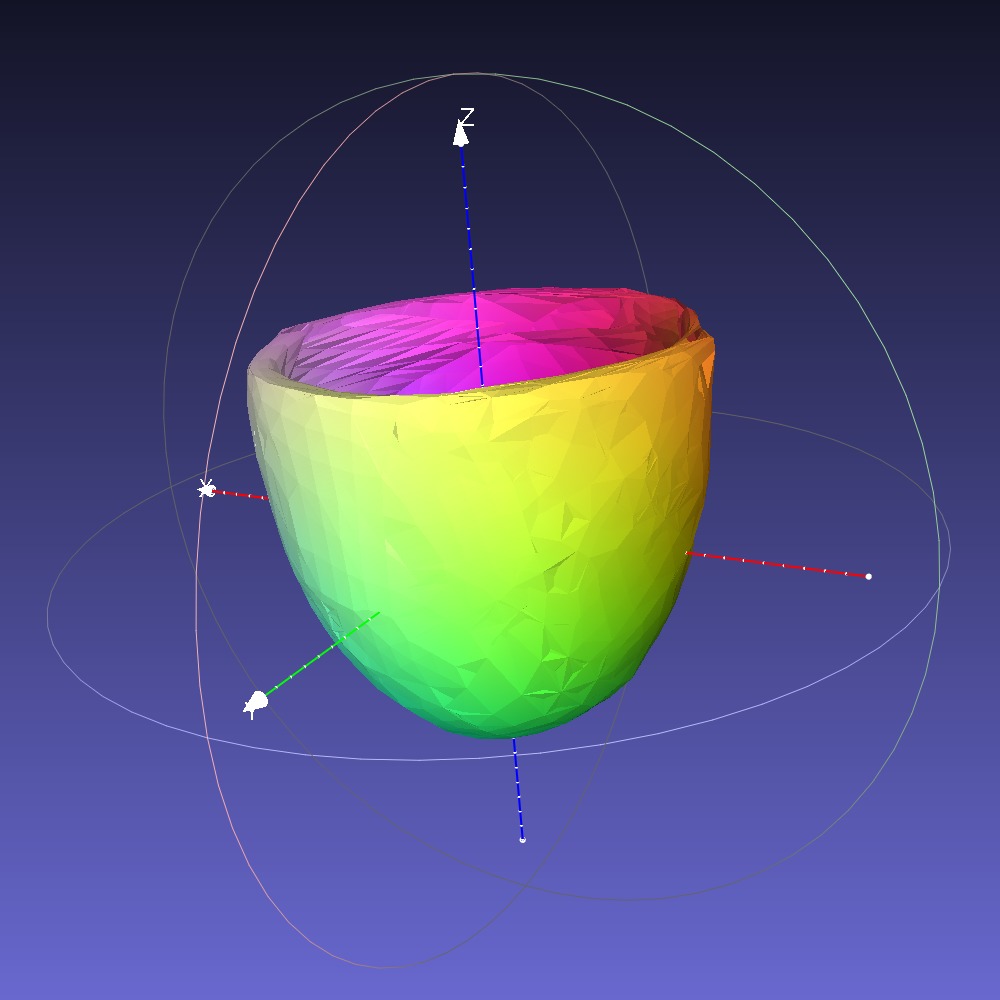}
    \end{tabular}
    \caption{\textbf{Qualitative results on synthetic validation dataset.} Left: Exemplary object pose estimations with rendered 3D bounding boxes, coordinate systems and shape meshes, overlaid on top of the respective input images. Centre: BEV images emphasizing accurate depth estimation of multiple objects in the scene (Our results are visualized in \emph{green} and ground truth in \emph{red}.). Right: Exemplary 3D mesh for each image, rendered with Meshlab \citep{meshlab}.}
    \label{fig:pose_comparison_synthetic}
\end{figure*}

Despite the monocular pose and shape estimation pipeline is the core focus of this work, it is worth mentioning that by employing ICP we can again strongly enhance performance. While we are a little worse in terms of 3D IoU at a threshold of 0.5 when comparing with NOCS \citep{wang2019normalized} and CASS\footnotemark[1] \citep{chen2020learning}, we are superior for the $5\degree\&5cm$ and $10\degree\&10cm$ metrics, despite no use of any real annotated data.
\footnotetext[1]{The numbers of CASS are different as in their paper since they used average precision instead. The authors provided us with their results for average recall.}

\begin{figure*}[t!]
    \centering
    \includegraphics[width=0.95\linewidth]{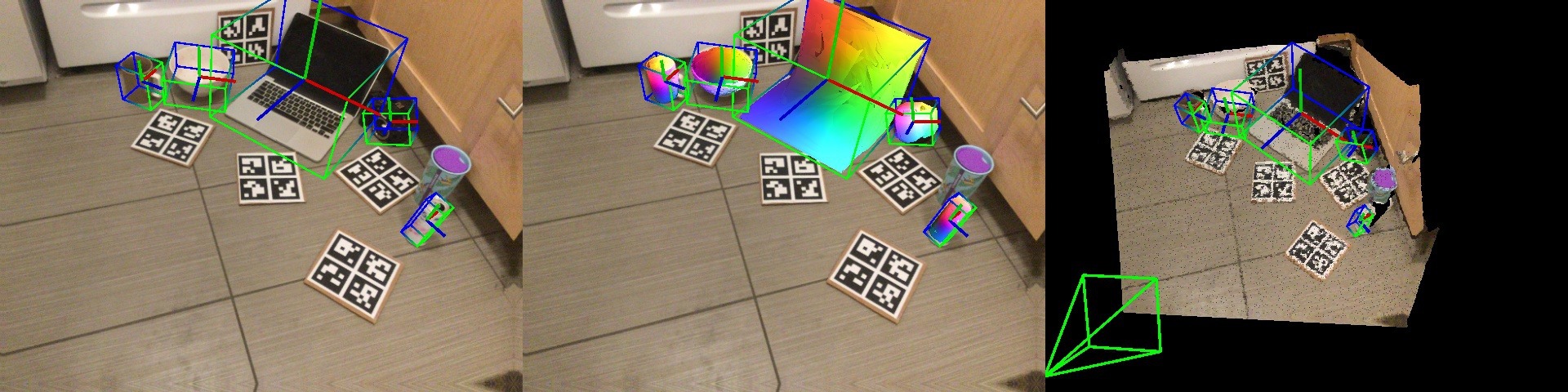}\\
    \includegraphics[width=0.95\linewidth]{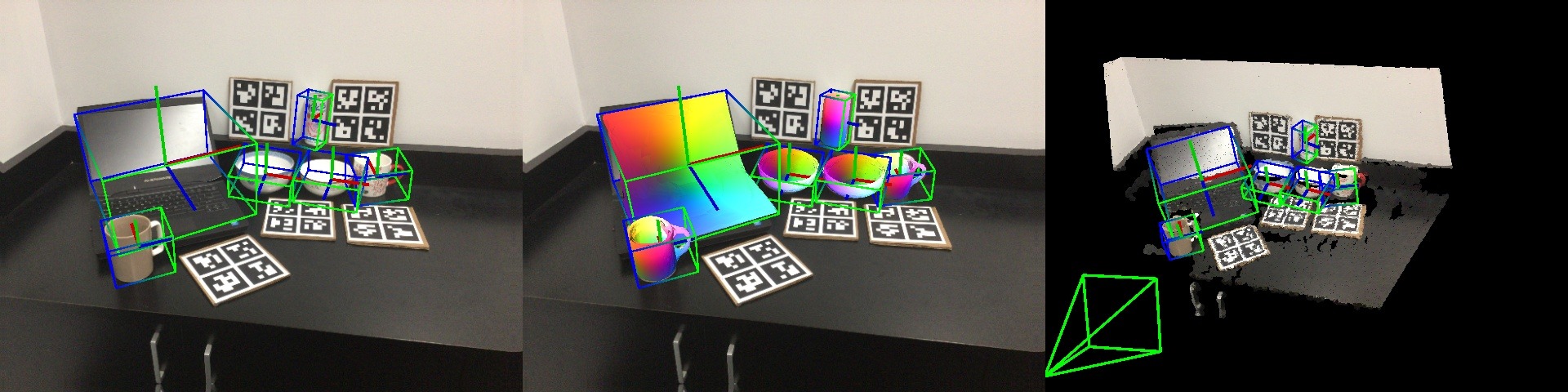}
     
    \caption{\textbf{Qualitative results on real test dataset.} Left: Object pose estimations with rendered 3D bounding boxes, coordinate systems (Centre:) and shape mesh, overlaid on top of the respective real input images. Right: We show an alternative viewpoint to the right to avoid ambiguities through projection.}
    \label{fig:pose_comparison_real}
\end{figure*}

Interestingly, as our objective function for self-super\-vision is highly motivated by the ICP formulation, we achieve almost the same results when running $\cps$ or $\cps$++ with ICP on real data.

\subsection{Qualitative Evaluation}

In Fig.~\ref{fig:pose_comparison_synthetic}, we show two qualitative examples for $\cps$. Notice how the extracted shapes, on the right, matches the perceived object in the scene. For instance, the long lens of the camera is properly reflected in the regressed 3D mesh. Additionally, the 3D bounding box overlap is high. Inspired by \citep{geiger2012we}, for better 3D understanding, we plotted the bird's-eye view visualization of the scene by conducting an orthographic projection on the X-Z plane (2nd from the left). Similarly, we computed an orthographic projection on the X-Y plane (3rd from left) as the ground plane assumption is invalid. We employ these projections to demonstrate that we can compute accurate 6D poses and 3D scales without being sensitive to ambiguities (scale \vs depth) due to monocular data. In fact, despite the ambiguity, our network is able to compute precise scales and poses as demonstrated in the tight overlaps.

Fig.~\ref{fig:pose_comparison_real} presents some qualitative results for $\cps$++ without ICP, demonstrating the models' capabilities for real applications.

\section{Conclusion}
In this paper we introduced $\cps$, the first method for monocular class-level 6D pose and shape estimation. We additionally proposed a novel point cloud alignment loss and experimentally demonstrated that it leads to performance that is on par or better than existing loss functions, while also favoring accurate reconstructions of the detected objects' geometry. As labeling data for the task at hand is very labor expensive, we trained our method purely on synthetic data, leaving a significant synthetic-to-real domain gap. Thus, to bridge the gap, we additionally tailored $\self$ towards the problem of class-level 6D pose estimation and recorded a RGB-D training dataset composed of over 30k frames, which we made publicly available. We demonstrate that leveraging our modified self-supervision leads to a significant leap forward when evaluating on real data, without relying on any annotations for real data.

\bibliographystyle{spbasic}      
{\footnotesize
\bibliography{shortstrings,vggroup,egbib}

\begin{thebibliography}{91}
\providecommand{\natexlab}[1]{#1}
\providecommand{\url}[1]{{#1}}
\providecommand{\urlprefix}{URL }
\expandafter\ifx\csname urlstyle\endcsname\relax
  \providecommand{\doi}[1]{DOI~\discretionary{}{}{}#1}\else
  \providecommand{\doi}{DOI~\discretionary{}{}{}\begingroup
  \urlstyle{rm}\Url}\fi
\providecommand{\eprint}[2][]{\url{#2}}

\bibitem[{Bernardini et~al.(1999)Bernardini, Mittleman, Rushmeier, Silva, and
  Taubin}]{bernardini1999ball}
Bernardini F, Mittleman J, Rushmeier H, Silva C, Taubin G (1999) The
  ball-pivoting algorithm for surface reconstruction. IEEE transactions on
  visualization and computer graphics 5(4):349--359

\bibitem[{Bousmalis et~al.(2017)Bousmalis, Silberman, Dohan, Erhan, and
  Krishnan}]{bousmalis2017unsupervisedPixelda}
Bousmalis K, Silberman N, Dohan D, Erhan D, Krishnan D (2017) Unsupervised
  pixel-level domain adaptation with generative adversarial networks. In: CVPR,
  pp 3722--3731

\bibitem[{Brachmann et~al.(2014)Brachmann, Krull, Michel, Gumhold, Shotton, and
  Rother}]{brachmann2014learning}
Brachmann E, Krull A, Michel F, Gumhold S, Shotton J, Rother C (2014) Learning
  6d object pose estimation using 3d object coordinates. In: ECCV, Springer, pp
  536--551

\bibitem[{Carr et~al.(2012)Carr, Sheikh, and Matthews}]{carr2012monocular}
Carr P, Sheikh Y, Matthews I (2012) Monocular object detection using 3d
  geometric primitives. In: European Conference on Computer Vision, Springer,
  pp 864--878

\bibitem[{Chang et~al.(2015)Chang, Funkhouser, Guibas, Hanrahan, Huang, Li,
  Savarese, Savva, Song, Su et~al.}]{chang2015shapenet}
Chang AX, Funkhouser T, Guibas L, Hanrahan P, Huang Q, Li Z, Savarese S, Savva
  M, Song S, Su H, et~al. (2015) Shapenet: An information-rich 3d model
  repository. arXiv preprint arXiv:151203012

\bibitem[{Chen et~al.(2020{\natexlab{a}})Chen, Li, Wang, and
  Xu}]{chen2020learning}
Chen D, Li J, Wang Z, Xu K (2020{\natexlab{a}}) Learning canonical shape space
  for category-level 6d object pose and size estimation. In: CVPR, pp
  11973--11982

\bibitem[{Chen et~al.(2019)Chen, Ling, Gao, Smith, Lehtinen, Jacobson, and
  Fidler}]{chen2019learning_dibrenderer}
Chen W, Ling H, Gao J, Smith E, Lehtinen J, Jacobson A, Fidler S (2019)
  Learning to predict 3d objects with an interpolation-based differentiable
  renderer. In: NeurIPS, pp 9605--9616

\bibitem[{Chen et~al.(2016)Chen, Kundu, Zhang, Ma, Fidler, and
  Urtasun}]{chen2016monocular}
Chen X, Kundu K, Zhang Z, Ma H, Fidler S, Urtasun R (2016) Monocular 3d object
  detection for autonomous driving. In: CVPR

\bibitem[{Chen et~al.(2017)Chen, Ma, Wan, Li, and Xia}]{chen2017multi}
Chen X, Ma H, Wan J, Li B, Xia T (2017) Multi-view 3d object detection network
  for autonomous driving. In: CVPR

\bibitem[{Chen et~al.(2020{\natexlab{b}})Chen, Tai, Sun, and
  Li}]{chen2020monopair}
Chen Y, Tai L, Sun K, Li M (2020{\natexlab{b}}) Monopair: Monocular 3d object
  detection using pairwise spatial relationships. In: CVPR, pp 12093--12102

\bibitem[{Cignoni et~al.(2008)Cignoni, Callieri, Corsini, Dellepiane,
  Ganovelli, and Ranzuglia}]{meshlab}
Cignoni P, Callieri M, Corsini M, Dellepiane M, Ganovelli F, Ranzuglia G (2008)
  {MeshLab: an Open-Source Mesh Processing Tool}. In: Eurographics Italian
  Chapter Conference, The Eurographics Association

\bibitem[{Deng et~al.(2020{\natexlab{a}})Deng, Genova, Yazdani, Bouaziz,
  Hinton, and Tagliasacchi}]{deng2020cvxnet}
Deng B, Genova K, Yazdani S, Bouaziz S, Hinton G, Tagliasacchi A
  (2020{\natexlab{a}}) Cvxnet: Learnable convex decomposition. In: CVPR, pp
  31--44

\bibitem[{Deng et~al.(2020{\natexlab{b}})Deng, Xiang, Mousavian, Eppner, Bretl,
  and Fox}]{deng2020self}
Deng X, Xiang Y, Mousavian A, Eppner C, Bretl T, Fox D (2020{\natexlab{b}})
  Self-supervised 6d object pose estimation for robot manipulation. In: ICRA

\bibitem[{Ding et~al.(2020)Ding, Huo, Yi, Wang, Shi, Lu, and
  Luo}]{ding2020learning}
Ding M, Huo Y, Yi H, Wang Z, Shi J, Lu Z, Luo P (2020) Learning depth-guided
  convolutions for monocular 3d object detection. In: CVPRW, pp 1000--1001

\bibitem[{Garon et~al.(2018)Garon, Laurendeau, and
  Lalonde}]{garon2018framework}
Garon M, Laurendeau D, Lalonde JF (2018) A framework for evaluating 6-dof
  object trackers. In: Proceedings of the European Conference on Computer
  Vision (ECCV), pp 582--597

\bibitem[{Geiger et~al.(2012)Geiger, Lenz, and Urtasun}]{geiger2012we}
Geiger A, Lenz P, Urtasun R (2012) Are we ready for autonomous driving? the
  kitti vision benchmark suite. In: CVPR

\bibitem[{Genova et~al.(2020)Genova, Cole, Sud, Sarna, and
  Funkhouser}]{genova2020local}
Genova K, Cole F, Sud A, Sarna A, Funkhouser T (2020) Local deep implicit
  functions for 3d shape. In: CVPR, pp 4857--4866

\bibitem[{Gkioxari et~al.(2019)Gkioxari, Malik, and Johnson}]{gkioxari2019mesh}
Gkioxari G, Malik J, Johnson J (2019) Mesh r-cnn. In: ICCV, pp 9785--9795

\bibitem[{Godard et~al.(2017)Godard, Mac~Aodha, and
  Brostow}]{godard2017unsupervised}
Godard C, Mac~Aodha O, Brostow GJ (2017) Unsupervised monocular depth
  estimation with left-right consistency. In: CVPR, pp 270--279

\bibitem[{Groueix et~al.(2018)Groueix, Fisher, Kim, Russell, and
  Aubry}]{groueix2018atlasnet}
Groueix T, Fisher M, Kim VG, Russell BC, Aubry M (2018) A papier-m{\^a}ch{\'e}
  approach to learning 3d surface generation. In: CVPR, pp 216--224

\bibitem[{He et~al.(2017)He, Gkioxari, Doll{\'a}r, and Girshick}]{he2017mask}
He K, Gkioxari G, Doll{\'a}r P, Girshick R (2017) Mask r-cnn. In: ICCV

\bibitem[{Hinterstoisser et~al.(2011)Hinterstoisser, Holzer, Cagniart, Ilic,
  Konolige, Navab, and Lepetit}]{Hinterstoisser2011}
Hinterstoisser S, Holzer S, Cagniart C, Ilic S, Konolige K, Navab N, Lepetit V
  (2011) {Multimodal templates for real-time detection of texture-less objects
  in heavily cluttered scenes}. In: ICCV

\bibitem[{Hinterstoisser et~al.(2012{\natexlab{a}})Hinterstoisser, Cagniart,
  Ilic, Sturm, Navab, Fua, and Lepetit}]{hinterstoisser2012gradient}
Hinterstoisser S, Cagniart C, Ilic S, Sturm P, Navab N, Fua P, Lepetit V
  (2012{\natexlab{a}}) Gradient response maps for real-time detection of
  textureless objects. TPAMI 34(5):876--888

\bibitem[{Hinterstoisser et~al.(2012{\natexlab{b}})Hinterstoisser, Lepetit,
  Ilic, Holzer, Bradski, Konolige, and Navab}]{Hinterstoisser2012}
Hinterstoisser S, Lepetit V, Ilic S, Holzer S, Bradski G, Konolige K, Navab N
  (2012{\natexlab{b}}) Model based training, detection and pose estimation of
  texture-less 3d objects in heavily cluttered scenes. In: ACCV, pp 548--562

\bibitem[{Hodan et~al.(2016)Hodan, Matas, and Obdrzalek}]{Hodan2016}
Hodan T, Matas J, Obdrzalek S (2016) {On Evaluation of 6D Object Pose
  Estimation}. In: ECCVW

\bibitem[{Hodan et~al.(2018)Hodan, Michel, Brachmann, Kehl, GlentBuch, Kraft,
  Drost, Vidal, Ihrke, Zabulis et~al.}]{hodan2018bop}
Hodan T, Michel F, Brachmann E, Kehl W, GlentBuch A, Kraft D, Drost B, Vidal J,
  Ihrke S, Zabulis X, et~al. (2018) Bop: Benchmark for 6d object pose
  estimation. In: ECCV

\bibitem[{Hoda{\v{n}} et~al.(2019)Hoda{\v{n}}, Vineet, Gal, Shalev, Hanzelka,
  Connell, Urbina, Sinha, and Guenter}]{hodan2019photorealistic}
Hoda{\v{n}} T, Vineet V, Gal R, Shalev E, Hanzelka J, Connell T, Urbina P,
  Sinha S, Guenter B (2019) Photorealistic image synthesis for object instance
  detection. ICIP

\bibitem[{Hodan et~al.(2020)Hodan, Barath, and Matas}]{hodan2020epos}
Hodan T, Barath D, Matas J (2020) Epos: Estimating 6d pose of objects with
  symmetries. In: CVPR, pp 11703--11712

\bibitem[{Hu et~al.(2019)Hu, Hugonot, Fua, and Salzmann}]{hu2019segpose}
Hu Y, Hugonot J, Fua P, Salzmann M (2019) Segmentation-driven 6d object pose
  estimation. In: CVPR, pp 3385--3394

\bibitem[{Jiang et~al.(2019)Jiang, Hou, Cao, Cheng, Wei, and
  Xiong}]{jiang2019integral}
Jiang PT, Hou Q, Cao Y, Cheng MM, Wei Y, Xiong HK (2019) Integral object mining
  via online attention accumulation. In: ICCV, pp 2070--2079

\bibitem[{Kato et~al.(2018)Kato, Ushiku, and Harada}]{kato2018renderer}
Kato H, Ushiku Y, Harada T (2018) Neural 3d mesh renderer. In: CVPR, pp
  3907--3916

\bibitem[{Kato et~al.(2020)Kato, Beker, Morariu, Ando, Matsuoka, Kehl, and
  Gaidon}]{kato2020differentiable}
Kato H, Beker D, Morariu M, Ando T, Matsuoka T, Kehl W, Gaidon A (2020)
  Differentiable rendering: A survey. arXiv preprint arXiv:200612057

\bibitem[{Kehl et~al.(2017)Kehl, Manhardt, Tombari, Ilic, and
  Navab}]{kehl2017ssd}
Kehl W, Manhardt F, Tombari F, Ilic S, Navab N (2017) {SSD-6D}: Making
  rgb-based 3{D} detection and 6{D} pose estimation great again. In: CVPR, pp
  1521--1529

\bibitem[{Kendall et~al.(2018)Kendall, Gal, and Cipolla}]{kendall2018multi}
Kendall A, Gal Y, Cipolla R (2018) Multi-task learning using uncertainty to
  weigh losses for scene geometry and semantics. In: CVPR

\bibitem[{Kocabas et~al.(2019)Kocabas, Karagoz, and Akbas}]{kocabas2019self}
Kocabas M, Karagoz S, Akbas E (2019) Self-supervised learning of 3d human pose
  using multi-view geometry. In: CVPR, pp 1077--1086

\bibitem[{Kolesnikov et~al.(2019)Kolesnikov, Zhai, and
  Beyer}]{kolesnikov2019revisiting}
Kolesnikov A, Zhai X, Beyer L (2019) Revisiting self-supervised visual
  representation learning. In: CVPR, pp 1920--1929

\bibitem[{Krull et~al.(2015)Krull, Brachmann, Michel, Ying~Yang, Gumhold, and
  Rother}]{krull2015learning}
Krull A, Brachmann E, Michel F, Ying~Yang M, Gumhold S, Rother C (2015)
  Learning analysis-by-synthesis for 6{D} pose estimation in {RGB-D} images.
  In: ICCV, pp 954--962

\bibitem[{Ku et~al.(2018)Ku, Mozifian, Lee, Harakeh, and
  Waslander}]{ku2018joint}
Ku J, Mozifian M, Lee J, Harakeh A, Waslander SL (2018) Joint 3d proposal
  generation and object detection from view aggregation. In: IROS

\bibitem[{Ku et~al.(2019)Ku, Pon, and Waslander}]{ku2019monocular}
Ku J, Pon AD, Waslander SL (2019) Monocular 3d object detection leveraging
  accurate proposals and shape reconstruction. In: CVPR

\bibitem[{Kundu et~al.(2018)Kundu, Li, and Rehg}]{kundu20183d}
Kundu A, Li Y, Rehg JM (2018) 3d-rcnn: Instance-level 3d object reconstruction
  via render-and-compare. In: CVPR

\bibitem[{Lee et~al.(2018)Lee, Tseng, Huang, Singh, and Yang}]{lee2018diverse}
Lee HY, Tseng HY, Huang JB, Singh M, Yang MH (2018) Diverse image-to-image
  translation via disentangled representations. In: ECCV, pp 35--51

\bibitem[{Li et~al.(2019{\natexlab{a}})Li, Chen, and Shen}]{li2019stereo}
Li P, Chen X, Shen S (2019{\natexlab{a}}) Stereo r-cnn based 3d object
  detection for autonomous driving. In: CVPR

\bibitem[{Li et~al.(2019{\natexlab{b}})Li, Wang, Ji, Xiang, and
  Fox}]{li2019deepim}
Li Y, Wang G, Ji X, Xiang Y, Fox D (2019{\natexlab{b}}) {DeepIM}: Deep
  iterative matching for 6d pose estimation. IJCV pp 1--22

\bibitem[{Li et~al.(2019{\natexlab{c}})Li, Wang, and Ji}]{li2019cdpn}
Li Z, Wang G, Ji X (2019{\natexlab{c}}) {CDPN}: {C}oordinates-{B}ased
  {D}isentangled {P}ose {N}etwork for {R}eal-{T}ime {RGB}-{B}ased 6-{DoF}
  {O}bject {P}ose {E}stimation. In: ICCV, pp 7678--7687

\bibitem[{Lin et~al.(2014)Lin, Maire, Belongie, Hays, Perona, Ramanan,
  Doll{\'a}r, and Zitnick}]{coco_eccv14}
Lin TY, Maire M, Belongie S, Hays J, Perona P, Ramanan D, Doll{\'a}r P, Zitnick
  CL (2014) Microsoft coco: Common objects in context. In: ECCV, pp 740--755

\bibitem[{Lin et~al.(2017)Lin, Goyal, Girshick, He, and
  Doll{\'a}r}]{lin2017focal}
Lin TY, Goyal P, Girshick R, He K, Doll{\'a}r P (2017) Focal loss for dense
  object detection. In: ICCV

\bibitem[{Liu et~al.(2018)Liu, Lehman, Molino, Such, Frank, Sergeev, and
  Yosinski}]{liu2018intriguing}
Liu R, Lehman J, Molino P, Such FP, Frank E, Sergeev A, Yosinski J (2018) An
  intriguing failing of convolutional neural networks and the coordconv
  solution. In: NeurIPS

\bibitem[{Liu et~al.(2019)Liu, Li, Chen, and Li}]{liu2019softras}
Liu S, Li T, Chen W, Li H (2019) Soft rasterizer: A differentiable renderer for
  image-based 3d reasoning. ICCV pp 7708--7717

\bibitem[{Liu et~al.(2016)Liu, Anguelov, Erhan, Szegedy, Reed, Fu, and
  Berg}]{liu2016ssd}
Liu W, Anguelov D, Erhan D, Szegedy C, Reed S, Fu CY, Berg AC (2016) Ssd:
  Single shot multibox detector. In: ECCV

\bibitem[{Loper and Black(2014)}]{opendr_eccv14}
Loper MM, Black MJ (2014) {OpenDR}: An approximate differentiable renderer. In:
  ECCV, vol 8695, pp 154--169

\bibitem[{Lowe(1999)}]{lowe1999object}
Lowe DG (1999) Object recognition from local scale-invariant features. In:
  ICCV, vol~2, pp 1150--1157

\bibitem[{Ma et~al.(2019)Ma, Wang, Li, Ouyang, and Zhang}]{ma2019accurate}
Ma X, Wang Z, Li H, Ouyang W, Zhang P (2019) Accurate monocular 3d object
  detection via color-embedded 3d reconstruction for autonomous driving. In:
  ICCV

\bibitem[{Manhardt et~al.(2018)Manhardt, Kehl, Navab, and
  Tombari}]{manhardt2018deep}
Manhardt F, Kehl W, Navab N, Tombari F (2018) Deep model-based 6d pose
  refinement in rgb. In: ECCV

\bibitem[{Manhardt et~al.(2019{\natexlab{a}})Manhardt, Arroyo, Rupprecht,
  Busam, Birdal, Navab, and Tombari}]{manhardt2019ambiguity}
Manhardt F, Arroyo D, Rupprecht C, Busam B, Birdal T, Navab N, Tombari F
  (2019{\natexlab{a}}) Explaining the ambiguity of object detection and 6d pose
  from visual data. In: ICCV

\bibitem[{Manhardt et~al.(2019{\natexlab{b}})Manhardt, Kehl, and
  Gaidon}]{Manhardt2019}
Manhardt F, Kehl W, Gaidon A (2019{\natexlab{b}}) Roi-10d: Monocular lifting of
  2d detection to 6d pose and metric shape. In: CVPR

\bibitem[{Manhardt et~al.(2019{\natexlab{c}})Manhardt, Kehl, and
  Gaidon}]{manhardt2019roi}
Manhardt F, Kehl W, Gaidon A (2019{\natexlab{c}}) {ROI-10D}: Monocular lifting
  of 2d detection to 6d pose and metric shape. In: CVPR, pp 2069--2078

\bibitem[{Mescheder et~al.(2019)Mescheder, Oechsle, Niemeyer, Nowozin, and
  Geiger}]{mescheder2019occupancy}
Mescheder L, Oechsle M, Niemeyer M, Nowozin S, Geiger A (2019) Occupancy
  networks: Learning 3d reconstruction in function space. In: CVPR, pp
  4460--4470

\bibitem[{Mousavian et~al.(2017)Mousavian, Anguelov, Flynn, and
  Kosecka}]{mousavian20173d}
Mousavian A, Anguelov D, Flynn J, Kosecka J (2017) 3d bounding box estimation
  using deep learning and geometry. In: CVPR

\bibitem[{Nguyen-Phuoc et~al.(2018)Nguyen-Phuoc, Li, Balaban, and
  Yang}]{nguyen2018rendernet}
Nguyen-Phuoc TH, Li C, Balaban S, Yang Y (2018) Rendernet: A deep convolutional
  network for differentiable rendering from 3d shapes. In: Advances in Neural
  Information Processing Systems, pp 7891--7901

\bibitem[{Nie et~al.(2020)Nie, Han, Guo, Zheng, Chang, and
  Zhang}]{nie2020total3dunderstanding}
Nie Y, Han X, Guo S, Zheng Y, Chang J, Zhang JJ (2020) Total3dunderstanding:
  Joint layout, object pose and mesh reconstruction for indoor scenes from a
  single image. In: CVPR, pp 55--64

\bibitem[{Niemeyer et~al.(2020)Niemeyer, Mescheder, Oechsle, and
  Geiger}]{niemeyer2020differentiable}
Niemeyer M, Mescheder L, Oechsle M, Geiger A (2020) Differentiable volumetric
  rendering: Learning implicit 3d representations without 3d supervision. In:
  CVPR, pp 3504--3515

\bibitem[{Park et~al.(2019{\natexlab{a}})Park, Florence, Straub, Newcombe, and
  Lovegrove}]{Park_2019_CVPR}
Park JJ, Florence P, Straub J, Newcombe R, Lovegrove S (2019{\natexlab{a}})
  Deepsdf: Learning continuous signed distance functions for shape
  representation. In: CVPR

\bibitem[{Park et~al.(2019{\natexlab{b}})Park, Patten, and
  Vincze}]{park2019pix2pose}
Park K, Patten T, Vincze M (2019{\natexlab{b}}) Pix2pose: Pixel-wise coordinate
  regression of objects for 6d pose estimation. In: ICCV

\bibitem[{Park et~al.(2020)Park, Mousavian, Xiang, and
  Fox}]{park2020latentfusion}
Park K, Mousavian A, Xiang Y, Fox D (2020) Latentfusion: End-to-end
  differentiable reconstruction and rendering for unseen object pose
  estimation. In: CVPR, pp 10710--10719

\bibitem[{Paszke et~al.(2019)Paszke, Gross, Massa, Lerer, Bradbury, Chanan,
  Killeen, Lin, Gimelshein, Antiga et~al.}]{paszke2019pytorch}
Paszke A, Gross S, Massa F, Lerer A, Bradbury J, Chanan G, Killeen T, Lin Z,
  Gimelshein N, Antiga L, et~al. (2019) Pytorch: An imperative style,
  high-performance deep learning library. In: NeurIPS, pp 8026--8037

\bibitem[{Peng et~al.(2019)Peng, Liu, Huang, Zhou, and Bao}]{peng2019pvnet}
Peng S, Liu Y, Huang Q, Zhou X, Bao H (2019) Pvnet: Pixel-wise voting network
  for 6dof pose estimation. In: CVPR

\bibitem[{Qi et~al.(2017)Qi, Su, Mo, and Guibas}]{qi2017pointnet}
Qi CR, Su H, Mo K, Guibas LJ (2017) Pointnet: Deep learning on point sets for
  3d classification and segmentation. In: CVPR

\bibitem[{Rad and Lepetit(2017)}]{rad2017bb8}
Rad M, Lepetit V (2017) {BB8}: A scalable, accurate, robust to partial
  occlusion method for predicting the 3{D} poses of challenging objects without
  using depth. In: ICCV, pp 3828--3836

\bibitem[{Ren et~al.(2015)Ren, He, Girshick, and Sun}]{ren2015faster}
Ren S, He K, Girshick R, Sun J (2015) Faster r-cnn: Towards real-time object
  detection with region proposal networks. In: NeurIPS

\bibitem[{Romea et~al.(2011)Romea, Torres, and Srinivasa}]{Romea-2011-7355}
Romea AC, Torres MM, Srinivasa S (2011) The moped framework: Object recognition
  and pose estimation for manipulation. International Journal of Robotics
  Research 30(10):1284 -- 1306

\bibitem[{Simonelli et~al.(2019)Simonelli, Rota~Bulo, Porzi, Lopez-Antequera,
  and Kontschieder}]{Simonelli2019}
Simonelli A, Rota~Bulo S, Porzi L, Lopez-Antequera M, Kontschieder P (2019)
  Disentangling monocular 3d object detection. In: ICCV

\bibitem[{Song and Xiao(2016)}]{song2016deep}
Song S, Xiao J (2016) Deep sliding shapes for amodal 3d object detection in
  rgb-d images. In: CVPR

\bibitem[{Sundermeyer et~al.(2018{\natexlab{a}})Sundermeyer, Marton, Durner,
  Brucker, and Triebel}]{Sundermeyer_2018_ECCV}
Sundermeyer M, Marton ZC, Durner M, Brucker M, Triebel R (2018{\natexlab{a}})
  Implicit 3d orientation learning for 6d object detection from rgb images. In:
  ECCV

\bibitem[{Sundermeyer et~al.(2018{\natexlab{b}})Sundermeyer, Marton, Durner,
  Brucker, and Triebel}]{sundermeyer2018implicit}
Sundermeyer M, Marton ZC, Durner M, Brucker M, Triebel R (2018{\natexlab{b}})
  Implicit 3d orientation learning for 6d object detection from rgb images. In:
  ECCV, pp 699--715

\bibitem[{Sundermeyer et~al.(2020)Sundermeyer, Durner, Puang, Marton,
  Vaskevicius, Arras, and Triebel}]{sundermeyer2020multi}
Sundermeyer M, Durner M, Puang EY, Marton ZC, Vaskevicius N, Arras KO, Triebel
  R (2020) Multi-path learning for object pose estimation across domains. In:
  CVPR, pp 13916--13925

\bibitem[{Tekin et~al.(2018)Tekin, Sinha, and Fua}]{tekin2018real}
Tekin B, Sinha SN, Fua P (2018) Real-time seamless single shot 6d object pose
  prediction. In: CVPR

\bibitem[{Tian et~al.(2019)Tian, Shen, Chen, and He}]{tian2019fcos}
Tian Z, Shen C, Chen H, He T (2019) {FCOS}: Fully convolutional one-stage
  object detection. In: ICCV, pp 9627--9636

\bibitem[{Tremblay et~al.(2018)Tremblay, To, and
  Birchfield}]{tremblay2018falling}
Tremblay J, To T, Birchfield S (2018) Falling things: A synthetic dataset for
  3d object detection and pose estimation. In: CVPRW, pp 2038--2041

\bibitem[{Umeyama(1991)}]{umeyama1991least}
Umeyama S (1991) Least-squares estimation of transformation parameters between
  two point patterns. IEEE Transactions on Pattern Analysis \& Machine
  Intelligence pp 376--380

\bibitem[{Vidal et~al.(2018)Vidal, Lin, Llad{\'o}, and
  Mart{\'\i}}]{vidal2018method}
Vidal J, Lin CY, Llad{\'o} X, Mart{\'\i} R (2018) A method for 6d pose
  estimation of free-form rigid objects using point pair features on range
  data. Sensors 18(8):2678

\bibitem[{Wang et~al.(2020)Wang, Manhardt, Shao, Ji, Navab, and
  Tombari}]{wang2020self6d}
Wang G, Manhardt F, Shao J, Ji X, Navab N, Tombari F (2020) Self6d:
  Self-supervised monocular 6d object pose estimation. In: ECCV

\bibitem[{Wang et~al.(2019)Wang, Sridhar, Huang, Valentin, Song, and
  Guibas}]{wang2019normalized}
Wang H, Sridhar S, Huang J, Valentin J, Song S, Guibas LJ (2019) Normalized
  object coordinate space for category-level 6d object pose and size
  estimation. In: CVPR

\bibitem[{Wang et~al.(2018)Wang, Zhang, Li, Fu, Liu, and
  Jiang}]{wang2018pixel2mesh}
Wang N, Zhang Y, Li Z, Fu Y, Liu W, Jiang YG (2018) Pixel2mesh: Generating 3d
  mesh models from single rgb images. In: ECCV, pp 52--67

\bibitem[{Wu et~al.(2020)Wu, Sahoo, and Hoi}]{wu2020recent}
Wu X, Sahoo D, Hoi SC (2020) Recent advances in deep learning for object
  detection. Neurocomputing 396:39 -- 64

\bibitem[{Xiang et~al.(2018)Xiang, Schmidt, Narayanan, and
  Fox}]{xiang2017posecnn}
Xiang Y, Schmidt T, Narayanan V, Fox D (2018) {PoseCNN}: A convolutional neural
  network for 6{D} object pose estimation in cluttered scenes. RSS

\bibitem[{Xu and Chen(2018)}]{Xu2018}
Xu B, Chen Z (2018) Multi-level fusion based 3d object detection from monocular
  images. In: CVPR

\bibitem[{Yang et~al.(2018)Yang, Feng, Shen, and Tian}]{yang2018foldingnet}
Yang Y, Feng C, Shen Y, Tian D (2018) Foldingnet: Point cloud auto-encoder via
  deep grid deformation. In: CVPR, pp 206--215

\bibitem[{Yu et~al.(2018)Yu, Tanner, Venkatraman, and Dieter}]{Xiang2018}
Yu X, Tanner S, Venkatraman N, Dieter F (2018) Posecnn: A convolutional neural
  network for 6d object pose estimation in cluttered scenes. In: RSS

\bibitem[{Zakharov et~al.(2019{\natexlab{a}})Zakharov, Kehl, and
  Ilic}]{zakharov2019deceptionnet}
Zakharov S, Kehl W, Ilic S (2019{\natexlab{a}}) Deceptionnet: Network-driven
  domain randomization. In: ICCV, pp 532--541

\bibitem[{Zakharov et~al.(2019{\natexlab{b}})Zakharov, Shugurov, and
  Ilic}]{zakharov2019dpod}
Zakharov S, Shugurov I, Ilic S (2019{\natexlab{b}}) Dpod: Dense 6d pose object
  detector in rgb images. In: ICCV

\bibitem[{Zakharov et~al.(2020)Zakharov, Kehl, Bhargava, and
  Gaidon}]{Zakharov_2020_CVPR}
Zakharov S, Kehl W, Bhargava A, Gaidon A (2020) Autolabeling 3d objects with
  differentiable rendering of sdf shape priors. In: CVPR

\end{thebibliography}
}

\end{document}